\documentclass[sigconf]{acmart}
\copyrightyear{2024}
\acmYear{2024}
\setcopyright{acmlicensed}
\acmConference[CIKM '24] {Proceedings of the 33rd ACM International Conference on Information and Knowledge Management}{October 21--25, 2024}{Boise, ID, USA.}
\acmBooktitle{Proceedings of the 33rd ACM International Conference on Information and Knowledge Management (CIKM '24), October 21--25, 2024, Boise, ID, USA}
\acmISBN{979-8-4007-0436-9/24/10}
\acmDOI{10.1145/3627673.3679582}
\usepackage{enumitem}
\usepackage{bm}
\usepackage{makecell}
\usepackage{multirow}
\usepackage{multicol}
\usepackage{amsfonts}
\usepackage{graphicx}
\usepackage{tabularx}
\usepackage{subfigure}
\newcommand{\modelname}{\texttt{EMERGE}}

\AtBeginDocument{%
  \providecommand\BibTeX{{%
    \normalfont B\kern-0.5em{\scshape i\kern-0.25em b}\kern-0.8em\TeX}}}

\begin{document}

\title{\modelname{}: Enhancing Multimodal Electronic Health Records Predictive Modeling with Retrieval-Augmented Generation}

\settopmatter{authorsperrow=4}
\author{Yinghao Zhu}
\authornote{Equal contribution.}
\orcid{0000-0002-2640-6477}
\affiliation{%
  \institution{Beihang University\\Peking University}
  \city{Beijing}
  \country{China}
}
\email{yhzhu99@gmail.com}

\author{Changyu Ren}
\authornotemark[1]
\orcid{0009-0004-6477-5160}
\affiliation{%
  \institution{Beihang University}
  \city{Beijing}
  \country{China}
}
\email{cyren@buaa.edu.cn}

\author{Zixiang Wang}
\orcid{0009-0000-1257-9580}
\affiliation{%
  \institution{Peking University}
  \city{Beijing}
  \country{China}
}
\email{wangzx@stu.pku.edu.cn}

\author{Xiaochen Zheng}
\orcid{0009-0007-9714-2193}
\affiliation{%
  \institution{ETH Zürich}
  \city{Zürich}
  \country{Switzerland}
}
\email{xzheng@ethz.ch}

\author{Shiyun Xie}
\orcid{0000-0001-5921-4060}
\affiliation{%
  \institution{Beihang University}
  \city{Beijing}
  \country{China}
}
\email{xieshiyun@buaa.edu.cn}

\author{Junlan Feng}
\orcid{0000-0001-5292-2945}
\affiliation{%
  \institution{China Mobile Research Institute}
  \city{Beijing}
  \country{China}
}
\email{fengjunlan@chinamobile.com}

\author{Xi Zhu}
\orcid{0009-0001-0222-066X}
\affiliation{%
  \institution{China Mobile Research Institute}
  \city{Beijing}
  \country{China}
}
\email{zhuqian@chinamobile.com}

\author{Zhoujun Li}
\orcid{0000-0002-9603-9713}
\affiliation{%
  \institution{Beihang University}
  \city{Beijing}
  \country{China}
}
\email{lizj@buaa.edu.cn}

\author{Liantao Ma}
\orcid{0000-0001-5233-0624}
\affiliation{%
  \institution{Peking University}
  \city{Beijing}
  \country{China}
}
\email{malt@pku.edu.cn}

\author{Chengwei Pan}
\authornote{Corresponding author.}
\orcid{0000-0003-0497-7903}
\affiliation{%
  \institution{Beihang University \\ Zhongguancun Laboratory}
  \city{Beijing}
  \country{China}
}
\email{pancw@buaa.edu.cn}

\renewcommand{\shortauthors}{Yinghao Zhu et al.}
\renewcommand{\shorttitle}{\modelname{}: Enhancing Multimodal EHR Predictive Modeling with RAG}

\begin{abstract}

The integration of multimodal Electronic Health Records (EHR) data has significantly advanced clinical predictive capabilities. Existing models, which utilize clinical notes and multivariate time-series EHR data, often fall short of incorporating the necessary medical context for accurate clinical tasks, while previous approaches with knowledge graphs (KGs) primarily focus on structured knowledge extraction. In response, we propose \modelname{}, a Retrieval-Augmented Generation (RAG) driven framework to enhance multimodal EHR predictive modeling. We extract entities from both time-series data and clinical notes by prompting Large Language Models (LLMs) and align them with professional PrimeKG, ensuring consistency. In addition to triplet relationships, we incorporate entities' definitions and descriptions for richer semantics. The extracted knowledge is then used to generate task-relevant summaries of patients' health statuses. Finally, we fuse the summary with other modalities using an adaptive multimodal fusion network with cross-attention. Extensive experiments on the MIMIC-III and MIMIC-IV datasets' in-hospital mortality and 30-day readmission tasks demonstrate the superior performance of the \modelname{} framework over baseline models. Comprehensive ablation studies and analysis highlight the efficacy of each designed module and robustness to data sparsity. \modelname{} contributes to refining the utilization of multimodal EHR data in healthcare, bridging the gap with nuanced medical contexts essential for informed clinical predictions. We have publicly released the code at \url{https://github.com/yhzhu99/EMERGE}.

\end{abstract}



\maketitle

\section{Introduction}

The advent of Electronic Health Records (EHR) marks a pivotal advancement in the way patient data is gathered and analyzed, contributing to a more effective and informed healthcare delivery system for clinical prediction~\cite{gao2024comprehensive,ma2023aicare,liao2024learnable}. This advancement is largely attributed to the utilization of multimodal EHR data, which primarily includes clinical notes and multivariate time-series data from patient records~\cite{zhang2022m3care,wang2024recentEHRsurvey,zhang2023improving}. Such data types are integral to healthcare prediction tasks, mirroring the holistic approach practitioners adopt by leveraging various patient data points to inform their clinical decisions and treatment strategies, rather than depending on a single data source~\cite{xyx2023vecocare}. Deep learning-based methods have become the mainstream approach, processing multimodal data to learn a mapping from heterogeneous inputs to output labels~\cite{choi2017gram,ma2018kame,zhang2022m3care}. However, in contrast to healthcare professionals, who have a deep understanding of medical contexts through extensive experience and knowledge, neural networks trained from scratch lack these insights into medical concepts~\cite{miotto2018deep}. Without deliberate integration of external knowledge, these networks often lack the ability or sensitivity to recognize crucial disease entities or laboratory test results within the EHR, essential for accurate prediction tasks~\cite{zhu2024larger}. In response, some recent studies have begun incorporating knowledge graphs to infuse additional medical insights into their analyses~\cite{ye2021medpath,gao2022medml}. These graphs offer a supplementary layer of clinically relevant concepts, thereby enhancing the model's ability to provide contextually meaningful representations and interpretable evidence~\cite{yang2023kerprint}. Despite these advancements, significant limitations remain in fully linking external knowledge with multiple EHR modalities, underscoring the imperative need for continuous research to integrate multi-source insights and improve the multimodal EHR data predictive modeling.

Previous methods integrating external medical knowledge into EHR data analysis tend to extract knowledge from data modalities such as ICD disease codes, patient conditions, procedures, and drugs, neglecting the use of clinical notes and time-series data, which are more common and practical~\cite{rajkomar2018scalable} (\textbf{Limitation 1}). Additionally, these methods primarily extract hierarchical and structured knowledge from clinical-context knowledge graphs. However, these medical concepts—entity names and their relationships into a graph have limited direct contribution to predictive tasks (\textbf{Limitation 2}). With Large Language Models (LLMs) like GPT-4~\cite{openai2023gpt4} demonstrating strong capabilities in diverse clinical tasks~\cite{zhu2024larger,wornow2023shaky,shi2024reslora} and serving as large medical knowledge graphs (KGs)~\cite{sun2023head}. By prompting the LLM, GraphCare~\cite{jiang2023graphcare} constructs a GPT-KG using structured condition, procedure, and drug record data, represented as triples (entity 1, relationship, entity 2). It further employs graph neural networks for downstream tasks. However, this approach encounters the hallucination issue~\cite{zhang2023hallucinationInLLM}, where LLMs may generate incorrect or fabricated information. To mitigate this, GraphCare collaborates with medical professionals to scrutinize and remove potentially harmful content, a process that is both complex and labor-intensive, requiring significant expertise to validate and refine the generated triples. Moreover, directly generating the KG via LLMs introduces a domain gap since this task is likely untrained for the LLMs, leading to potentially lower accuracy compared to professional knowledge graphs built through established methodologies (\textbf{Limitation 3}).

To overcome these limitations, we propose utilizing LLMs in a Retrieval-augmented Generation (RAG) approach~\cite{lewis2020rag}. The RAG framework integrates structured time-series EHR data, unstructured clinical notes, and an established KG (PrimeKG~\cite{chandak2023buildingPrimeKG}) with LLM's semantic reasoning capabilities~\cite{wang2023can}. The LLMs are prompted to generate comprehensive summaries of patients' health statuses, and these summaries are then fused for downstream tasks. Despite its apparent simplicity, applying this method to clinical tasks presents several technical challenges:

\textbf{Challenge 1: How to extract entities from multimodal EHR data and match these entities with external KG consistently?} Extracting entities from the diverse and complex formats of EHR data (including clinical notes and multivariate time-series data) is challenging. Moreover, unlike structured codes where it can directly compare the code-related entities' embedding with KG's entity, the entities extracted by LLM have hallucination issues. Accurately matching extracted entities with those in an external knowledge graph while eliminating the potential for hallucinations posed by LLMs is crucial for maintaining the integrity and reliability of the clinical prediction tasks~\cite{imrie2023redefining}.

\textbf{Challenge 2: How to encode and incorporate long-text retrieved knowledge with task-relevant characteristics?} The extracted textual knowledge likely contains too many tokens~\cite{xiao2023efficient} for conventional language model inputs (e.g., BERT supports only 512 tokens~\cite{devlin2018bert}). However, with the development of long-context LLMs~\cite{zhu2023ExtendContextWindow}, it is feasible to leverage LLMs to distill this knowledge further. Additionally, simply integrating the retrieved knowledge may not be task-specific, creating a gap between the knowledge and downstream tasks~\cite{bai2023GripRank,bai2024infusing,bai2023kinet}. Therefore, a task-relevant prompting strategy~\cite{openai_prompt_engineering} is necessary during the LLM distillation process.

To these ends, We propose \modelname{} framework to address the above limitations and challenges with the following approaches, which are our three-fold contributions:
\begin{enumerate}[leftmargin=*,topsep=0pt]
    \item We design a RAG-driven multimodal EHR enhancement framework for clinical notes and time-series EHR data (\textbf{Response to Limitation 1}). \modelname{} leverages the capabilities of LLMs and professionally labeled large medical knowledge graphs. We retrieve medical entities by prompting the LLM for clinical notes and using z-score-based filtering for time-series data, then match them in KG with post-validation and alignment to mitigate hallucination (\textbf{Response to Limitation 3}). In addition to triples of entities, we also include more knowledge by extending the entities' definition and description. (\textbf{Response to Limitation 2}).
    \item Methodologically, we first compare LLM-generated entities with original clinical notes to ensure the entities appear in the raw text. We then compute their embeddings and cosine similarities among extracted entities and KG entities, aligning the entities through threshold-based filtering. This ensures that the overall entity extraction and matching process adheres to clinical standards with consistency guarantees (\textbf{Response to Challenge 1}). We prompt the long-context LLM to summarize the extracted knowledge into a distilled reflection of the patient's health status, instructing the generated content is task-relevant. To integrate the extracted knowledge and consider heterogeneity, we design an adaptive multimodal fusion network with a cross-attention mechanism that attentively fuses each modality's representation (\textbf{Response to Challenge 2}).
    \item Experimentally, our extensive experiments on the MIMIC-III and MIMIC-IV datasets, focusing on in-hospital mortality and 30-day readmission tasks, demonstrate \modelname{}'s superior performance and the effectiveness of each designed module. Additionally, to meet practical clinical needs, we evaluate the model's robustness with fewer training samples, showing \modelname{}'s remarkable resilience against data sparsity.
\end{enumerate}

\section{Related Work}

\subsection{Multimodal EHR Learning}

Advances in medical technology enable analysis of various medical modalities, including clinical notes, time-series lab data, demographics, conditions, procedures, drugs, and imaging. MedGTX~\cite{park2022MedGTX} introduces a pre-trained model for joint multi-modal representation learning, interpreting structured data as a graph and using a graph-text multi-modal framework. M3Care~\cite{zhang2022m3care} addresses missing modalities by imputing task-related information in the latent space with auxiliary data from similar patients, employing a modality-adaptive similarity metric to handle missing data. ~\citet{zhang2023improving} explore irregular time intervals in time-series EHR data and clinical notes via a time attention mechanism. ~\citet{xyx2023vecocare} propose a joint learning approach from visit sequences and clinical notes, using Gromov-Wasserstein Distance for contrastive learning and dual-channel retrieval to enhance patient similarity analysis. ~\citet{lee2023learning} introduce a unified framework for learning across all EHR modalities with modality-aware attention mechanisms, avoiding separate imputation modules.

Despite their effectiveness, these methods often overlook clinical background information, where external medical knowledge could enhance EHR data insights. The absence of semantic medical knowledge also complicates the training pipeline, especially with limited data.

\subsection{Incorporating External Knowledge for EHR}

To integrate clinical knowledge with EHR data, several studies leverage medical knowledge graphs (KGs) to enhance EHR representation learning and predictive performance. GRAM~\cite{choi2017gram} uses hierarchical medical ontologies via a graph attention network to refine medical representations. KAME~\cite{ma2018kame} embeds ontology information throughout the prediction process, enriching contextual understanding. MedPath~\cite{ye2021medpath} employs graph neural networks to integrate high-order connections from KGs into input representations. MedRetriever~\cite{ye2021MedRetriever} enhances health risk prediction and interpretability by combining EHR embeddings with features from disease-specific documents. Collaborative graph learning models like CGL~\cite{ijcai2021CGL} explore patient-disease interactions and domain knowledge, while KerPrint~\cite{yang2023kerprint} addresses knowledge decay across multiple visits. Recent advancements in Large Language Models (LLMs) as comprehensive knowledge bases~\cite{sun2023head} offer new possibilities, as seen in GraphCare~\cite{jiang2023graphcare}, which creates a KG from structured EHR data for GNN learning, despite challenges like hallucination.

These studies primarily focus on structured medical data, often neglecting the rich semantic information in unstructured EHR data. This limitation underscores the need for methods that comprehensively utilize both structured and unstructured data.

\section{Problem Formulation}

\subsection{EHR Datasets Formulation}

The electronic health records (EHR) dataset comprises both structured and unstructured data, represented as multivariate time-series data and clinical notes, respectively. To facilitate analysis, these two modalities are initially processed separately, either from the raw data matrix or via a tokenization process. Specifically, the multivariate time-series data, denoted as $\bm{x}_{TS} \in \mathbb{R}^{T \times F}$, encapsulate information across $T$ visits and $F$ numeric or categorical features. Clinical notes, denoted as $\bm{x}_{Note}$, contain recorded notes documenting the health status of each patient. Additionally, external knowledge graphs (KGs) are incorporated to enhance the personalized representation of each patient.

\subsection{Predictive Objective Formulation}

The prediction objective is conceptualized as a binary classification task, which involves predicting in-hospital mortality and 30-day readmission. By leveraging the comprehensive patient information derived from EHR data and KGs, the model aims to predict specific clinical outcomes. The prediction task is formulated as:
\begin{equation}
\hat{y} = \text{Framework}(\bm{x}_{TS}, \bm{x}_{Note}, KG) 
\end{equation}
where $\hat{y}$ represents the targeted prediction outcome.

For the in-hospital mortality prediction task, our objective is to determine the discharge status based on data from the initial 48-hour window of an ICU stay, where a status of 0 indicates the patient is alive and 1 indicates the patient is deceased. In the same vein, the 30-day readmission task aims to predict whether a patient will be readmitted within 30 days after discharge, with 0 indicating no readmission and 1 indicating readmission.

\section{Methodology}

\begin{figure*}[!ht]
  \centering
  \includegraphics[width=0.9\linewidth]{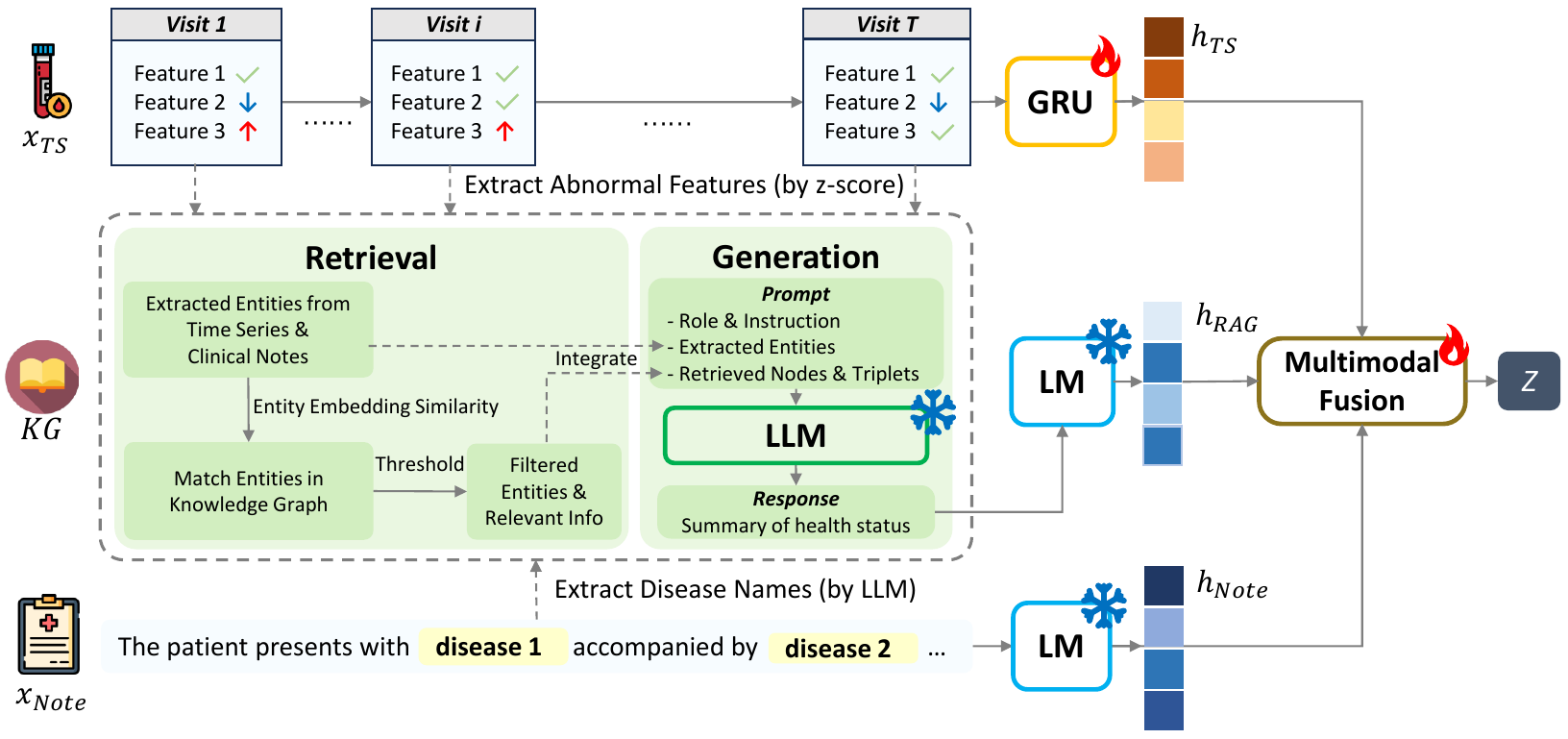}
  \caption{\textit{Overall architecture of our proposed \modelname{} framework.} The modules enclosed within the dashed box illustrate the RAG-driven enhancement pipeline. ``LM'' denotes Language Model (basically BERT-based model), while ``LLM'' in this paper normally refers to the GPT-based Large Language Model.}
  \label{fig:overall_pipeline}
\end{figure*}

Figure~\ref{fig:overall_pipeline} shows the overall framework architecture of \modelname{}.

\subsection{Multimodal EHR Embedding Extraction}

We delve into the techniques used for embedding extraction from multimodal EHR, emphasizing the transformation from raw, human-readable inputs, denoted as $\bm{x}$, to deep semantic embeddings $\bm{h}$ for comprehensive analysis.

When dealing with time-series data, we employ the Gated Recurrent Unit (GRU) network as the encoder. GRU is a highly efficient variant of recurrent neural networks, capable of capturing the time dependencies in sequence data and encoding this temporally linked information. We extract the representation of time-series data as follows:
\begin{equation}
\bm{h}_{TS} = {\rm GRU}(x_{TS}),
\end{equation}
where $x_{TS}$ is the time-series data and $h_{TS}$ denotes the output of the time-series encoder.

As for text records, we utilize a medical domain language model to obtain text embeddings, represented as $\text{TextEncoder}$: 
\begin{equation}
\bm{h}_{Note} = {\rm TextEncoder}(\bm{x}_{Note}).
\end{equation}
where $\bm{x}_{Note}$ are the textual clinical notes and ${h}_{Note}$ denotes the note representation.

\subsection{RAG-Driven Enhancement Pipeline}

\subsubsection{Extract Entities from Multimodal EHR Data}

To exploit the expert information encapsulated within the knowledge graph, it is necessary to extract disease entities from both time-series data and clinical notes, and subsequently align them with the information present in the graph. The set of disease entities in the time-series data is denoted as $\bm{E}_{TS}$, while those in the clinical notes data are denoted as $\bm{E}_{Note}$. Naturally, we design two separate processes tailored to each modality.

\begin{figure}[!ht]
    \centering
    \includegraphics[width=0.8\linewidth]{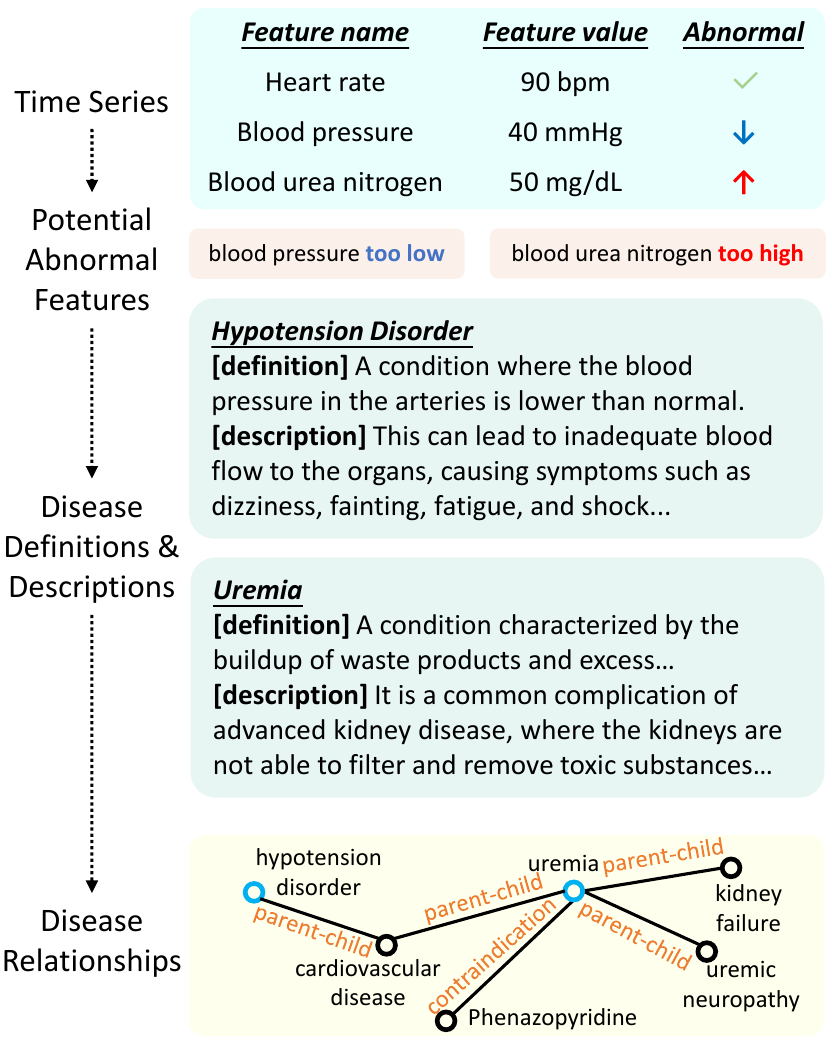}
    \caption{\textit{Process of information retrieval for time-series data.}}
    \label{fig:ts_rag}
\end{figure}

\paragraph{Retrieval process for time-series data.} Time-series data is a structured format encompassing feature names and resultant values post-clinical examination. Each feature name reflects specific aspects of an individual's physical condition, highlighting the deviations from the reference range. As shown in Figure~\ref{fig:ts_rag}, the specified record showcases low blood pressure and high blood urea nitrogen, significantly surpassing the normal range. This implies the potential risk of hypotension and uremia for the patient. Indeed, such feature names occur in disease definitions and descriptions, typically indicating serious health threats.

For each patient, there are usually more than one entity (or abnormal feature), and some may be missing values. Consequently, our focus is primarily on non-empty values. For each feature $\bm{x}_{TS_i}$, we can identify outliers through the z-score method~\cite{curtis2016mystery}, which measures anomalies by calculating the deviation of data points from the mean, using standard deviation as a unit as below:
\begin{equation}
s_i = \frac{\bm{x}_{TS_i}- {\rm mean}(\bm{x}_{TS_i})}{{\rm std}(\bm{x}_{TS_i})}
\end{equation}
where $s_i$ represents the z-score of the $i$-th feature of a patient. Features over a specified threshold $\epsilon$ (such as 3-$\sigma$ deviation) are identified as abnormal, indicating potential health issues.

\begin{figure}[!ht]
    \centering
    \includegraphics[width=0.8\linewidth]{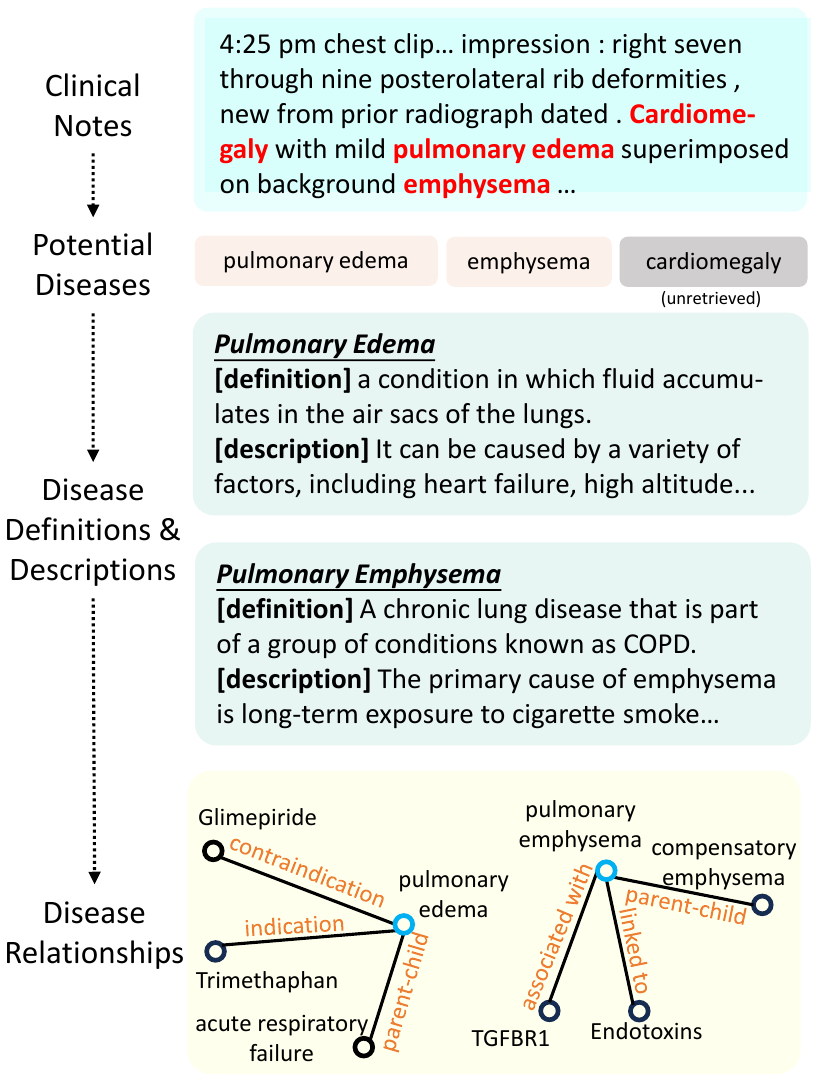}
    \caption{\textit{Process of information retrieval for textual clinical notes.} The grey block in potential diseases means no corresponding node found in external KG.}
    \label{fig:note_rag}
\end{figure}

\paragraph{Retrieval process for clinical notes.} Unlike structured data, clinical notes are presented in a textual format, which makes it challenging to comprehend and extract valuable information. However, LLMs have exhibited exceptional performance on natural language understanding tasks, including named entity recognition (NER). Therefore, we utilize an LLM to identify potential disease names that the patient may have, as shown in Figure~\ref{fig:note_rag}. Moreover, we implement specified rules for effective post-processing. 

\begin{figure}[!ht]
    \centering
    \includegraphics[width=1.0\linewidth]{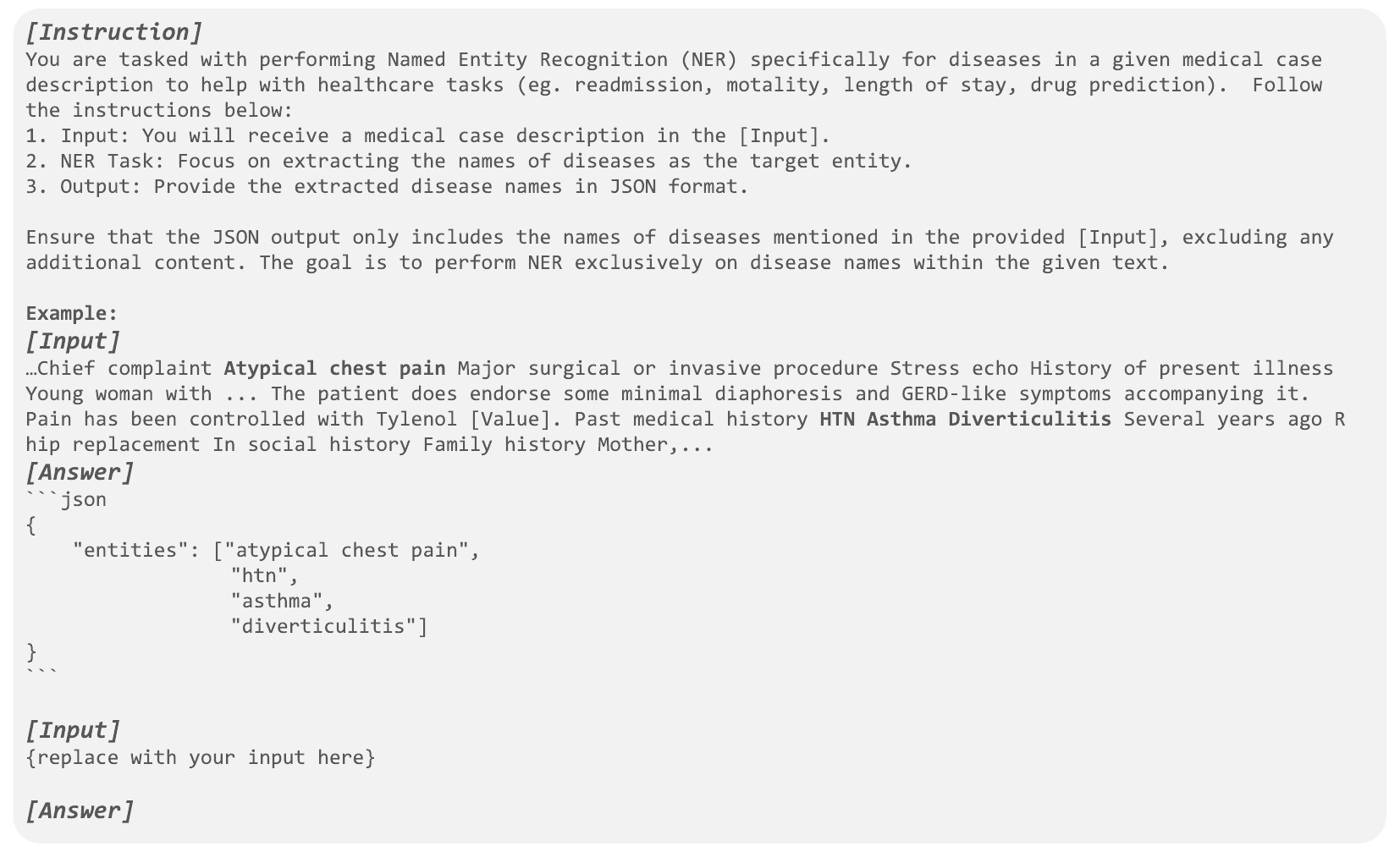}
    \caption{\textit{Prompt template for extracting entities.}}
    \label{fig:extract_entity_prompt}
\end{figure}
\begin{enumerate}[leftmargin=*]
\item \textbf{Entities Extraction:} We include an example and provide clear instructions in the prompt (Figure~\ref{fig:extract_entity_prompt}), instructing the LLM to concentrate on disease entities that the patient may suffer from. Sometimes, there may be no entities yielded in a single invocation, so we utilize multiple rounds to incrementally expand the current extracted entity set as shown below:
\begin{align}
 \bm{E}^i_{Note} &= {\rm LLM}(concat(\bm{P}_{Extract},\bm{x}_{Note})) \\
 \bm{E}_{Note} &\leftarrow \bm{E}_{Note} \bigcup \bm{E}^{i}_{Note}
\end{align}
where $\bm{P}_{extract}$ represents the prompt template. $\bm{E}^i_{Note}$ represents the entity set obtained in the $i$-th round and $\bm{E}_{Note}$ represents the aggregate set.

\item \textbf{Entities Refinement:} Considering the hallucination issue associated with LLMs, we design a post-processing process to address it. This process consists of three primary steps: first, we discard entities that do not appear in the original text; second, we leverage an LLM to filter entities not in the disease type; and finally, we delete duplicated entities to prevent semantic redundancy.
\begin{equation}
\bm{E}_{Note} \leftarrow \bm{E}_{Note} - \bm{E}_{illegal},
\end{equation}
where $\bm{E}_{illegal}$ denotes the illegal entity set, which we then remove from $\bm{E}_{Note}$.
\end{enumerate}

To ensure the quantity and quality of the extracted entities, we execute steps 1 and 2 iteratively until achieving convergence. 

\subsubsection{Retrieve Information from External KG}

To ensure an accurate match between the extracted entities and nodes within the knowledge graph, we adapt a semantic-based dense vector retrieval approach. Initially, we utilize a sentence embedding model denoted as $\text{TextEncoder}$ to encode all KG nodes, denoted as $Nodes$. Subsequently, for each entity in $\bm{E}_{TS}$ or $\bm{E}_{Note}$, we deploy the same embedding model to encode them. This process ensures that all embeddings are aligned within the same vector space, as shown below:
\begin{align}
    \bm{h}_{n}&= {\rm TextEncoder}(n), n \in Nodes \\
    \bm{h}_{e}&= {\rm TextEncoder}(e), e \in E
\end{align}
where $n$ and $e$ symbolize disease entities from $Nodes$ and the extracted entity set, respectively. ${h}_{n}$ and ${h}_{e}$ denote their corresponding embeddings.

When matching relative nodes, we take the current entity $e$ (including abnormal features and potential disease names) as the query. Then, we compute the similarities between $e$ and each node in the KG. The metric used for these calculations is cosine similarity:
\begin{equation}
\theta^{n}_{e} = \frac{{\bm{h}_n} \cdot {\bm{h}_e}}{\|{\bm{h}_n}\| \|{\bm{h}_e}\|}
\end{equation}
where $\bm{h}_{e}$ and $\bm{h}_{n}$ are embeddings of the entity $e$ and the node.

We establish a threshold to gauge the requisite similarity between two embeddings. We focus on nodes that surpass this threshold, ensuring that only the most relevant matches are considered:
\begin{equation}
f(e, \text{Nodes}) = 
\begin{cases}
\{ \hat{n} \} & \text{if } \theta^{\hat{n}}_{e} > \eta, \\
\emptyset & \text{otherwise},
\end{cases}
\end{equation}
where $\hat{n} = \arg\max_{n \in \text{Nodes}} \theta^{n}_{e}$, $\eta$ is the threshold for similarity, and $f(e, \text{Nodes})$ denotes the set of nodes that we exclusively accept as matches for the entity $e$.

Subsequently, we can obtain the definitions and descriptions within the disease entities, each represented as a node of the graph. Furthermore, relationships between diseases, encapsulated within triples, act as the edges of the graph. These pieces of information elaborate on the severity of the diseases, the harm they pose to the human body, and their interconnections from various perspectives. They further clarify the entity information from the original notes, thereby enhancing the LLM's understanding of the patient's health condition.

\subsubsection{Summarize and Encode KG Knowledge}
Drawing from the entities extracted from time-series data and clinical notes, along with supplementary information about them, we have compiled extensive details about the patient's medical condition. However, this content contains too many tokens for conventional language model inputs (such as BERT). As a countermeasure, we utilize retrieval-augmented generation to condense the aforementioned details, thereby attaining a concise representation of the patient's health status. 

\begin{figure}[!ht]
    \centering
    \includegraphics[width=0.8\linewidth]{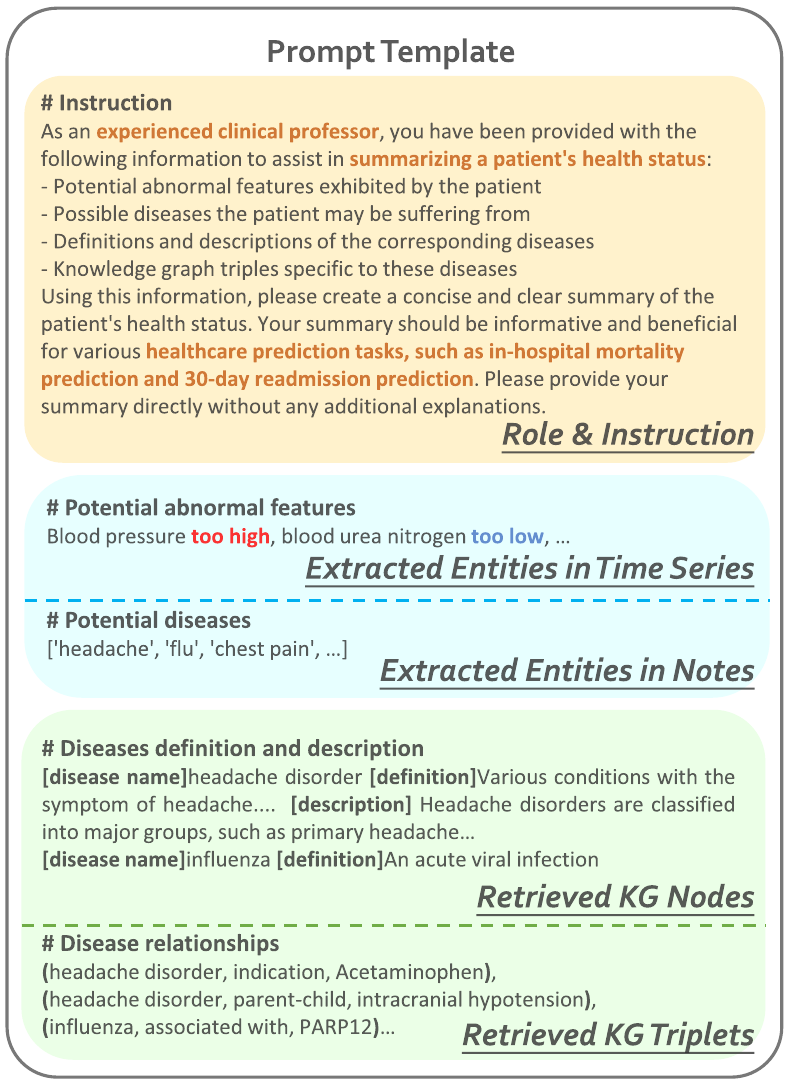}
    \caption{\textit{Prompt template for summary generation.}}
    \label{fig:prompt_template}
\end{figure}

The prompt template, as illustrated in Figure~\ref{fig:prompt_template}, begins by defining a role and instructions to guide the generation by the LLM. Subsequently, we enumerate all abnormal features derived from the time-series data, and disease names extracted from clinical notes, which reflect the patient's health threats. To enhance comprehension, we integrate retrieved disease definitions and descriptions, along with the relationships sampled from the KG to form a comprehensive supplementary resource. Based on this augmented information, the LLM compiles a summary of the patient's health status.

Finally, we employ a language model, denoted as $\text{TextEncoder}$, to encode the retrieved knowledge from the external KG as below:
\begin{equation}
\bm{h}_{RAG} = {\rm TextEncoder}(\bm{x}_{RAG})
\end{equation}
where $\bm{h}_{RAG}$ symbolizes the sentence embedding of the summary, which we will combine with $\bm{h}_{TS}$ and $\bm{h}_{Note}$ to obtain a comprehensive representation of the patient's health status.

\subsection{Multimodal Fusion Network}

\begin{figure}[!ht]
    \centering
    \includegraphics[width=0.8\linewidth]{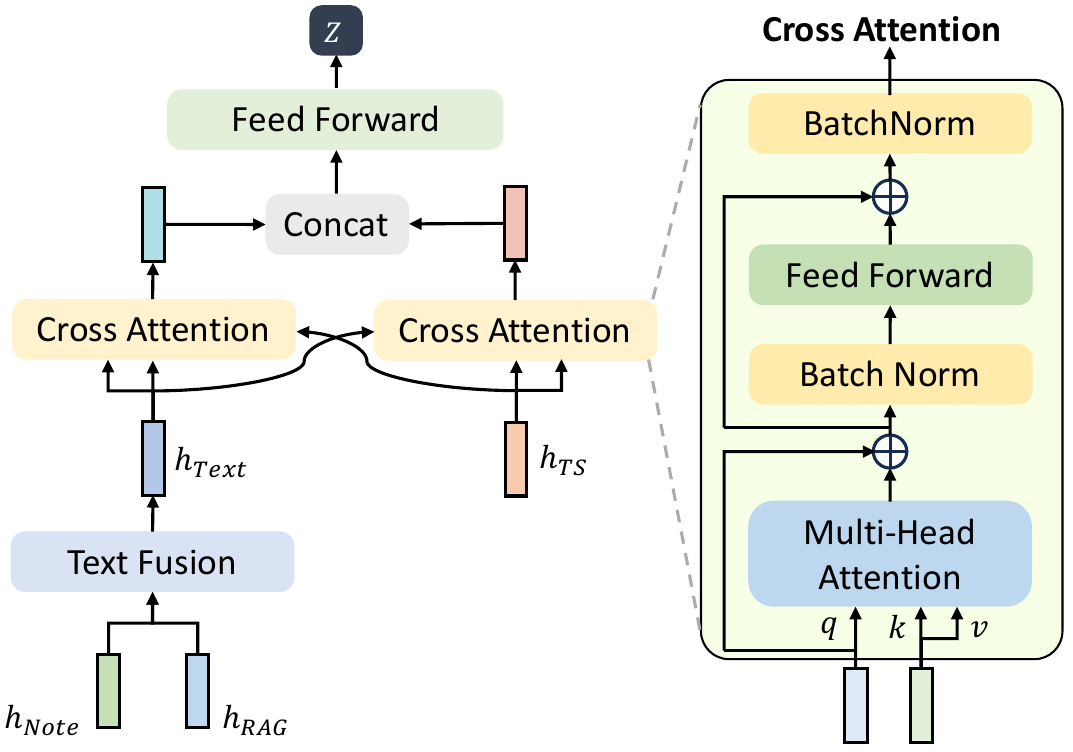}
    \caption{\textit{Fusion module.} It combines multimodal embeddings with attention mechanism into a fused representation.}
    \label{fig:fusion_module}
\end{figure}

Currently, there are three learned hidden representations, denoted respectively as $\bm{h}_{TS}$, $\bm{h}_{Note}$, and $\bm{h}_{RAG}$. 
We first concatenate the hidden representations extracted from entities with those from the text, and then utilize a fusion network to combine and map them to a unified dimension:
\begin{equation}
\begin{aligned}
    \bm{h}_{Text}={\rm TextFusion}({\rm Concat}\left[\bm{h}_{Note}, \bm{h}_{RAG}\right])
\end{aligned}
\end{equation}

To better integrate information from different modalities, we propose an attention-based fusion network primarily consisting of cross-attention layers. First, the $Query$ vector is computed from the hidden representation of the other modality, while the $Key$ and $Value$ vectors are computed from the hidden representations of the current modality:
\begin{equation}
\begin{aligned}
    {Q}_{Text} = \bm{W}_{q}\cdot {h}_{Text}&, \ \  {Q}_{TS} = \bm{W}_{q}\cdot {h}_{TS} \\ 
    {K}_{TS} = \bm{W}_{k}\cdot {h}_{TS}&, \ \  {K}_{Text} = \bm{W}_{k}\cdot {h}_{Text} \\ 
    {V}_{TS} = \bm{W}_{v}\cdot {h}_{TS}&, \ \  {V}_{Text} = \bm{W}_{v}\cdot {V}_{Text} \\ 
\end{aligned}
\end{equation}
where $Q$, $K$, $V$ are the $Query$, $Key$, $Value$ vectors respectively, and $\bm{W}_{q}$, $\bm{W}_{k}$, $\bm{W}_{v}$ are the corresponding projection matrices. Following this, we compute the attention outputs as follows:
\begin{equation}
\begin{aligned}
    \bm{z}_{Text}&={\rm softmax}(\frac{{Q}_{TS}  \bm{{K}}_{Text}^\top}{\sqrt{d_k}}) \cdot {V}_{Text} \\ 
    \bm{z}_{TS}&={\rm softmax}(\frac{{Q}_{Text} \bm{{K}}_{TS}^\top}{\sqrt{d_k}}) \cdot {V}_{TS} \\
\end{aligned}
\end{equation}

In addition, we apply residual connections and BatchNorm to every multi-head attention layer and feedforward network.

As a result, the outputs of the two cross-attention modules have carried information from both modalities. We further concatenate them and use an MLP layer to obtain the fused information:
\begin{equation}
\begin{aligned}
    \bm{z}&={\rm MLP}({\rm Concat}\left[\bm{z}_{TS}, \bm{z}_{Text}\right])
\end{aligned}
\end{equation}

Finally, the fused representation $\bm{z}$ is expected to predict downstream tasks. We pass $\bm{z}$ through a two-layer MLP structure, with an additional dropout layer between two fully connected layers, to obtain the final prediction results $\hat{y}$:
\begin{equation}
\begin{aligned}
    \hat{y}={\rm MLP}(\bm{z})
\end{aligned}
\end{equation}

The BCE Loss is selected as the loss function for the binary mortality outcome and readmission prediction task:
\begin{equation}
    \begin{aligned}
    \mathcal{L}(\hat{y}, y) = -\frac{1}{N}\sum_{i=1}^{N}(y_i \log(\hat{y}_i) + (1 - y_i) \log(1 - \hat{y}_i)) 
    \end{aligned}
\end{equation}
where $N$ is the number of patients within one batch, $\hat{y}\in [0,1]$ is the predicted probability, and $y$ is the ground truth.

By converting these three different types of data into compatible embeddings, our model lays a solid groundwork for the multimodal analysis of EHR. This strategy of embedding extraction sets the stage for further analysis tasks under the RAG framework, allowing us to accurately and comprehensively understand and analyze the complex information in EHR.

\section{Experimental Setups}

\subsection{Experimented Datasets and Utilized KG}

Sourced from the EHRs of the Beth Israel Deaconess Medical Center, MIMIC-III and MIMIC-IV dataset is extensive and widely used in healthcare research. We adhere to the established EHR benchmark pipeline~\cite{gao2024comprehensive,zhu2024pyehr} for preprocessing time-series data. 17 lab test features (include categorical features) and 2 demographic features (age and gender) are extracted. To minimize missing data, we consolidate every consecutive 12-hour segment into a single record for each patient, focusing on the first 48 records. And we follow Clinical-LongFormer\cite{li2023ClinicalLongFormer}'s approach to extract and preprocess clinical notes, which includes minimal but essential steps: removing all de-identification placeholders to protect Protected Health Information (PHI), replacing non-alphanumeric characters and punctuation marks, converting all letters to lowercase for consistency, and stripping extra white spaces.

We excluded all patients without any notes or time-series data. We randomly split the dataset into training, validation, and test set with 7:1:2 percentage. The statistics of datasets is in Table~\ref{tab:statistics_datasets}.

\begin{table}[!ht]
    \footnotesize
    \centering
    \caption{\textit{Statistics of datasets after preprocessing.} The number and proportion for labels are the percentage of the label with value $1$. $Out.$ denotes Mortality Outcome, $Re.$ denotes Readmission.}
    \label{tab:statistics_datasets}
    \begin{tabular}{ccccc}
        \toprule
        \textbf{Dataset} & \textbf{Split} & \textbf{Samples} & \textbf{$\text{Label}_{Out.}$} & \textbf{$\text{Label}_{Re.}$} \\
        \midrule
        \multirow{3}{*}{MIMIC-III} & Train & 10776 (70.00\%) & 1389 (12.89\%) & 1787 (16.58\%) \\
                                   & Val   & 1539 (10.00\%) & 193 (12.54\%) & 258 (16.76\%) \\
                                   & Test  & 3080 (20.00\%) & 361 (11.72\%) & 489 (15.88\%) \\
        \midrule
        \multirow{3}{*}{MIMIC-IV}  & Train & 13531 (70.00\%) & 1608 (11.88\%) & 2099 (15.51\%) \\
                                   & Val   & 1933 (10.00\%) & 244 (12.62\%) & 297 (15.36\%) \\
                                   & Test  & 3867 (20.00\%) & 448 (11.59\%) & 599 (15.49\%) \\
        \bottomrule
    \end{tabular}
\end{table}

The external knowledge base we utilized is PrimeKG~\cite{chandak2023buildingPrimeKG}, which integrates 20 high-quality resources to describe 17,080 diseases with 4,050,249 relationships representing ten major biological scales, including disease-associated entities. Futhermore, PrimeKG extracts textual features of disease nodes containing information about disease prevalence, symptoms, etiology, risk factors, epidemiology, clinical descriptions, management and treatment, complications, prevention, and when to seek medical attention, which are highly relevant to the clinical prediction tasks.

The median number of retrieved entities is 14 for MIMIC-III and 7 for MIMIC-IV, with an average effective extracted entity rate of 67.25\% and 66.88\%, respectively, from a total of 468,948 and 319,893 extracted entities for the two datasets.

When prompting the LLM to generate the summary, 4 patients in the MIMIC-III dataset were not successfully generated due to DeepSeek-v2~\cite{deepseekai2024deepseekv2}'s strict content censor policy, which flagged ``Content Exists Risk.'' We replaced these with ``None''.

\begin{table*}[!ht]
\footnotesize
\centering
\caption{\textit{In-hospital mortality and 30-day readmission prediction results on the MIMIC-III and MIMIC-IV datasets.} \textbf{Bold} indicates the best performance. All metrics are multiplied by 100 for readability purposes.}
\label{tab:overall_performance_table}
\resizebox{\textwidth}{!}{
\begin{tabular}{c|ccc|ccc|ccc|ccc}
\toprule
\multicolumn{1}{c|}{\multirow{2}{*}{\textbf{Methods}}} & \multicolumn{3}{c|}{\textbf{MIMIC-III Mortality}} & \multicolumn{3}{c|}{\textbf{MIMIC-III Readmission}} & \multicolumn{3}{c|}{\textbf{MIMIC-IV Mortality}} & \multicolumn{3}{c}{\textbf{MIMIC-IV Readmission}}\\
& AUROC ($\uparrow$)      & AUPRC ($\uparrow$)      & min(+P, Se) ($\uparrow$) 
& AUROC ($\uparrow$)      & AUPRC ($\uparrow$)      & min(+P, Se) ($\uparrow$)  
& AUROC ($\uparrow$)      & AUPRC ($\uparrow$)      & min(+P, Se) ($\uparrow$) 
& AUROC ($\uparrow$)      & AUPRC ($\uparrow$)      & min(+P, Se) ($\uparrow$)   \\
\midrule
MPIM  & 85.24±1.12 & 50.52±2.56 & 50.59±2.33  &  78.65±1.51 & 48.26±2.84 & 46.94±1.97  & 89.45±0.59 & 60.10±1.67 & 57.62±1.41 & 79.13±0.78 & 47.67±1.95 & 49.52±1.99  \\
UMM  & {84.01±1.10} & {49.76±2.21} & {49.41±2.45}  &  {77.46±1.36} & {47.81±2.55} & {47.27±1.91}  & 87.82±0.73 & 53.84±2.35 & 55.40±1.98 & 78.75±0.63 & 48.63±1.45 & 49.58±1.29 \\
MedGTX & 85.97±1.04 & 49.36±3.05 & 48.20±2.27 & 78.60±1.17 & 46.44±2.69 & 45.99±2.60 & 88.77±0.73 & 58.33±2.31 & 58.25±1.59 & 78.82±1.32 & 47.48±1.88 & 49.54±1.76 \\
VecoCare & {83.43±1.49} & {47.28±2.68} & {47.92±2.22}  &  {76.93±1.82} & {46.18±2.76} & {47.22±2.63}  & 88.01±0.68 & 55.37±2.20 & 55.35±1.72 & 79.17±1.20 & 51.58±1.93 & 51.42±1.48 \\
M3Care & {83.33±1.24} & {47.86±2.33} & {49.96±1.99}  &  {76.80±1.55} & {46.29±2.62} & {45.38±2.32}  & 88.14±0.78 & 54.06±2.04 & 54.30±1.73 & 79.87±1.31 & 51.03±1.95 & 51.10±1.36 \\
GRAM & {84.70±1.34} & {49.21±4.45} & {49.64±2.85}  &  {77.84±1.49} & {47.97±3.68} & {46.95±2.12}  & 87.75±0.65 & 54.01±2.93 & 54.62±2.63 & 79.53±1.01 & 50.13±2.53 & 50.80±1.67 \\
KAME & {84.59±1.11} & {49.48±3.37} & {49.51±2.33}  &  {78.04±1.34} & {48.23±3.21} & {47.41±2.50}  & 87.76±0.67 & 55.74±2.37 & 54.79±1.44 & 78.91±1.01 & 47.62±1.66 & 49.63±1.28 \\
CGL & {84.20±1.16} & {47.64±3.47} & {47.67±2.61}  &  {77.47±1.33} & {46.68±3.33} & {47.73±2.25}  & 88.42±0.94 & 56.64±2.21 & 54.80±1.62 & 78.95±0.90 & 47.74±1.66 & 49.16±1.24 \\
KerPrint & {85.29±1.21} & {51.23±3.48} & {50.88±2.24}  &  78.81±1.68 & 47.92±2.45 & 47.32±2.52  & 88.28±0.60 & 57.90±1.80 & 55.12±1.46 & 79.84±1.03 & 53.55±1.61 & 52.34±1.64 \\
MedPath & 85.61±1.34 & 48.90±3.24 & 48.86±3.00 & 77.92±0.85 & 45.66±2.61 & 45.72±2.24 & 88.85±1.00 & 56.82±2.60 & 57.96±2.63 & 78.88±0.83 & 47.58±2.23 & 49.75±2.39 \\
MedRetriever & 85.62±1.47 & 49.99±3.06 & 49.03±2.54 & 77.77±0.90 & 46.81±2.36 & 46.89±2.08 &  89.01±0.42 & 57.75±1.60 & 58.16±1.32 & 79.15±0.90 & 48.26±1.08 & 49.49±1.18 \\ 
GraphCare & 85.85±0.95 & 50.16±2.20 & 49.15±2.57 & 78.70±1.19 & 47.19±2.33 & 46.82±2.04 &  89.13±0.57 & 60.85±2.01 & 59.16±1.85 & 79.18±1.15 & 48.55±1.86 & 49.64±1.58  \\
\midrule
\modelname{} & \textbf{86.25±1.50} & \textbf{52.08±2.87} & \textbf{51.42±2.40} & \textbf{79.06±1.05} & \textbf{48.59±2.52} & \textbf{47.86±2.58} & \textbf{89.50±0.57} & \textbf{63.11±2.12} & \textbf{59.95±1.49} & \textbf{80.61±1.09} & \textbf{57.28±2.01} &  \textbf{54.50±1.71} \\
\bottomrule
\end{tabular}
}
\end{table*}

\subsection{Evaluation Metrics}

We adopt the following evaluation metrics, which are widely used in binary classification tasks:

\begin{itemize}[leftmargin=*]
    \item \textbf{AUROC}: This metric is our primary consideration in binary classification tasks due to its widespread use in clinical settings and its effectiveness in handling imbalanced datasets~\cite{auroc_better}.
    \item \textbf{AUPRC}: The AUPRC is particularly useful for evaluating performance in datasets with a significant imbalance between classes~\cite{kim2022auprc}.
    \item \textbf{min(+P, Se)}: This composite metric represents the minimum value between precision (+P) and sensitivity (Se), providing a balanced measure of model performance~\cite{ma2022safari}.
\end{itemize}

All these three metrics are the higher the better.

\subsection{Baseline Models}

\subsubsection{EHR Prediction Models} We include multimodal EHR baseline models (MPIM~\cite{zhang2023improving}, UMM~\cite{lee2023learning}, MedGTX~\cite{park2022MedGTX}, VecoCare\cite{xyx2023vecocare}, M3Care~\cite{zhang2022m3care}) and approaches that incorporating external knowledge from KG (GRAM~\cite{choi2017gram}, KAME~\cite{ma2018kame}, CGL~\cite{ijcai2021CGL}, KerPrint~\cite{yang2023kerprint}, MedPath~\cite{ye2021medpath}, MedRetriever~\cite{ye2021MedRetriever}), and LLM facilitated model GraphCare~\cite{jiang2023graphcare} as our baselines.

\subsubsection{Multimodal Fusion Methods} To examine the effectiveness of our fusion network, we consider fusion methods: Add~\cite{wu2018multi}, Concat~\cite{khadanga2019using, deznabi2021predicting}, Tensor Fusion (TF)~\cite{zadeh2017tensor}, and MAG~\cite{rahman2020integrating, yang2021leverage}.

\subsection{Implementation Details}

\subsubsection{Hardware and Software Configuration}
All runs are trained on a single Nvidia RTX 3090 GPU with CUDA 12.4. The server's system memory (RAM) size is 128GB. We implement the model in Python 3.11, PyTorch 2.2.2~\cite{paszke2019pytorch}, PyTorch Lightning 2.2.4~\cite{falcon2019lightning}, and pyehr~\cite{zhu2024pyehr}.

\subsubsection{Model Training and Hyperparameters}
AdamW~\cite{loshchilov2017decoupled} is employed with a batch size of 256 patients. All models are trained for 100 epochs with an early stopping strategy based on AUPRC after 10 epochs without improvement. The learning rate ${0.01,0.001,0.0001}$ and hidden dimensions ${32, 64, 128, 256}$ are tuned using a grid search strategy on the validation set. The searched hyperparameter for \modelname{} is: 128 hidden dimensions, 0.001 learning rate. The dropout rate is set to $0.25$. Performance is reported in the form of mean±std by applying bootstrapping on all test set samples 10 times for the MIMIC-III and MIMIC-IV datasets, following practices in \citet{ma2023aicare}. The threshold \(\epsilon\) for identifying anomalies in time-series data is set as 2 (z-score value=2). The threshold \(\eta\) for matching entities in KG is set as 0.6 for MIMIC-III and 0.7 for MIMIC-IV.

\subsubsection{Utilized (Large) Language Models}

\modelname{} utilizes both Language Models (LMs) and Large Language Models (LLMs) in the pipeline. For LMs, we use the frozen-parameter pretrained Clinical-LongFormer~\cite{li2023ClinicalLongFormer}'s [CLS] token~\cite{devlin2018bert} for extracting textual embeddings and BGE-M3~\cite{bge-m3} as the text embedding model to compute entity embeddings. For LLMs, we deploy an offline Qwen-7B~\cite{qwen} to extract entities from clinical notes and call the DeepSeek-V2 Chat~\cite{deepseekai2024deepseekv2} API to generate summaries.

\section{Experimental Results and Analysis}

\subsection{Experimental Results}

The performance of our \modelname{} framework on in-hospital mortality and 30-day readmission prediction tasks on the MIMIC-III and MIMIC-IV datasets is summarized in Table~\ref{tab:overall_performance_table}. \modelname{} consistently outperforms the baseline models, indicating its superior practical applicability in real-world clinical settings.

\subsection{Ablation Studies}

\subsubsection{Comparing Different Modality Fusion Strategies}

To understand the contribution of each modality and the modality fusion approaches, we compare their performance, as illustrated in Table~\ref{tab:ablation_performance_modality_fusion}. The results reveal that: 1) Utilizing multiple modalities is better than using a single modality. 2) The RAG pipeline-generated summary exhibits stronger representation capability (by comparing the settings ``Note only'' vs. ``RAG only'', and ``TS+Note'' vs. ``TS+RAG''). This showcases the effectiveness of task-relevant generated summaries in facilitating prediction modeling. 3) \modelname{}'s cross-attention-based adaptive multimodal fusion network outperforms other modality fusion strategies.

\begin{table*}[!ht]
\centering
\footnotesize
\caption{\textit{Ablation study results of 1) comparing each modality with RAG enhancement, and 2) comparing different multimodal fusion networks.} \textbf{Bold} and \underline{Underlined} indicates the best and 2nd best performance. All metrics are multiplied by 100.}
\label{tab:ablation_performance_modality_fusion}
\resizebox{\textwidth}{!}{
\begin{tabular}{l|ccc|ccc|ccc|ccc}
\toprule
\multicolumn{1}{c|}{\multirow{2}{*}{\textbf{Methods}}} & \multicolumn{3}{c|}{\textbf{MIMIC-III Mortality}}       & \multicolumn{3}{c|}{\textbf{MIMIC-III Readmission}}     & \multicolumn{3}{c|}{\textbf{MIMIC-IV Mortality}}        & \multicolumn{3}{c}{\textbf{MIMIC-IV Readmission}}      \\ 
\multicolumn{1}{c|}{}                                  & AUROC ($\uparrow$)   & AUPRC ($\uparrow$)   & min(+P, Se) ($\uparrow$) & AUROC ($\uparrow$)   & AUPRC ($\uparrow$)   & min(+P, Se) ($\uparrow$) & AUROC ($\uparrow$)   & AUPRC ($\uparrow$)   & min(+P, Se) ($\uparrow$) & AUROC ($\uparrow$)   & AUPRC ($\uparrow$)   & min(+P, Se) ($\uparrow$) \\
\midrule
TS only   & 84.57±1.50 & 46.53±3.14 & 48.89±2.92 & 77.17±1.36 & 43.87±2.72 & 46.21±2.83  
          & 87.96±0.65 & 55.62±2.00 & 55.02±2.01 & 79.03±1.17 & 51.79±1.93 & 51.02±1.66 \\
Note only & 66.50±1.40 & 19.62±0.68 & 23.22±1.23 & 64.76±1.00 & 24.64±0.76 & 27.07±0.51  
          & 69.47±1.03 & 27.70±1.26 & 30.90±1.30 & 66.40±0.97 & 29.52±1.31 & 32.39±1.61 \\
RAG only  & 69.21±1.54 & 22.46±2.68 & 27.04±2.62 & 64.65±1.05 & 24.12±1.78 & 27.65±1.63  
          & 71.84±1.27 & 27.68±2.76 & 30.62±2.91 & 67.37±1.29 & 28.26±2.37 & 31.83±2.16 \\
TS+Note   & 85.72±1.34 & 49.02±2.76 & 48.28±2.36 & \underline{78.36±1.06} & \underline{46.95±2.49} & 45.79±2.17 
          & 88.55±0.58 & 60.01±1.84 & 57.95±1.47 & 79.93±0.94 & 54.29±1.67 & 52.84±1.45 \\
TS+RAG    & \underline{86.21±1.29} & \underline{51.15±3.24} & \underline{50.62±2.78} & 78.24±0.90 & 46.94±2.54 & \underline{47.11±2.46} 
          & 89.49±0.58 & 62.49±2.19 & 58.75±2.20 & 80.55±1.12 & 55.64±2.07 & 52.38±1.77 \\
Note+RAG  & 72.32±1.14 & 27.07±1.66 & 28.66±1.72 & 68.80±0.80 & 28.87±1.47 & 31.96±1.62  
          & 74.96±1.12 & 32.28±2.97 & 35.43±2.54 & 70.72±1.23 & 32.42±2.26 & 35.33±2.70 \\
\midrule
TS+Text: \texttt{Concat} & 85.66±1.44 & 49.41±2.89 & 48.18±3.09 & 78.04±1.00 & 46.72±2.36 & 46.18±2.21 & 89.33±0.57 & 62.42±2.10 & 59.75±1.23 & 80.58±0.96 & 55.40±1.84 & 52.77±1.47 \\
TS+Text: \texttt{TF}     & 85.55±1.42 & 50.30±2.92 & 50.11±3.24 & 77.83±1.15 & 46.73±2.50 & 46.70±2.59 & 89.08±0.57 & 59.47±2.28 & 59.53±1.53 & 80.34±0.96 & 53.01±1.87 & 51.81±1.35 \\
TS+Text: \texttt{MAG}    & 86.09±1.47 & 49.14±2.51 & 49.12±2.92 & 77.69±0.89 & 44.86±2.04 & 45.76±1.67 & \textbf{89.56±0.62} & \underline{62.64±2.04} & \textbf{60.16±1.52} & \textbf{80.66±1.08} & \underline{56.62±1.96} & \underline{53.97±1.71} \\
TS+Text: \texttt{Ours}   & \textbf{86.25±1.50} & \textbf{52.08±2.87} & \textbf{51.42±2.40} & \textbf{79.06±1.05} & \textbf{48.59±2.52} & \textbf{47.86±2.58} & \underline{89.50±0.57} & \textbf{63.11±2.12} & \underline{59.95±1.49} & \underline{80.61±1.09} & \textbf{57.28±2.01} &  \textbf{54.50±1.71} \\
\bottomrule
\end{tabular}
}
\end{table*}

\subsubsection{Comparing Different Time-series Encoders}

\begin{figure}[!ht]
\centering
\subfigure[MIMIC-III]{
  \includegraphics[width=0.45\linewidth]{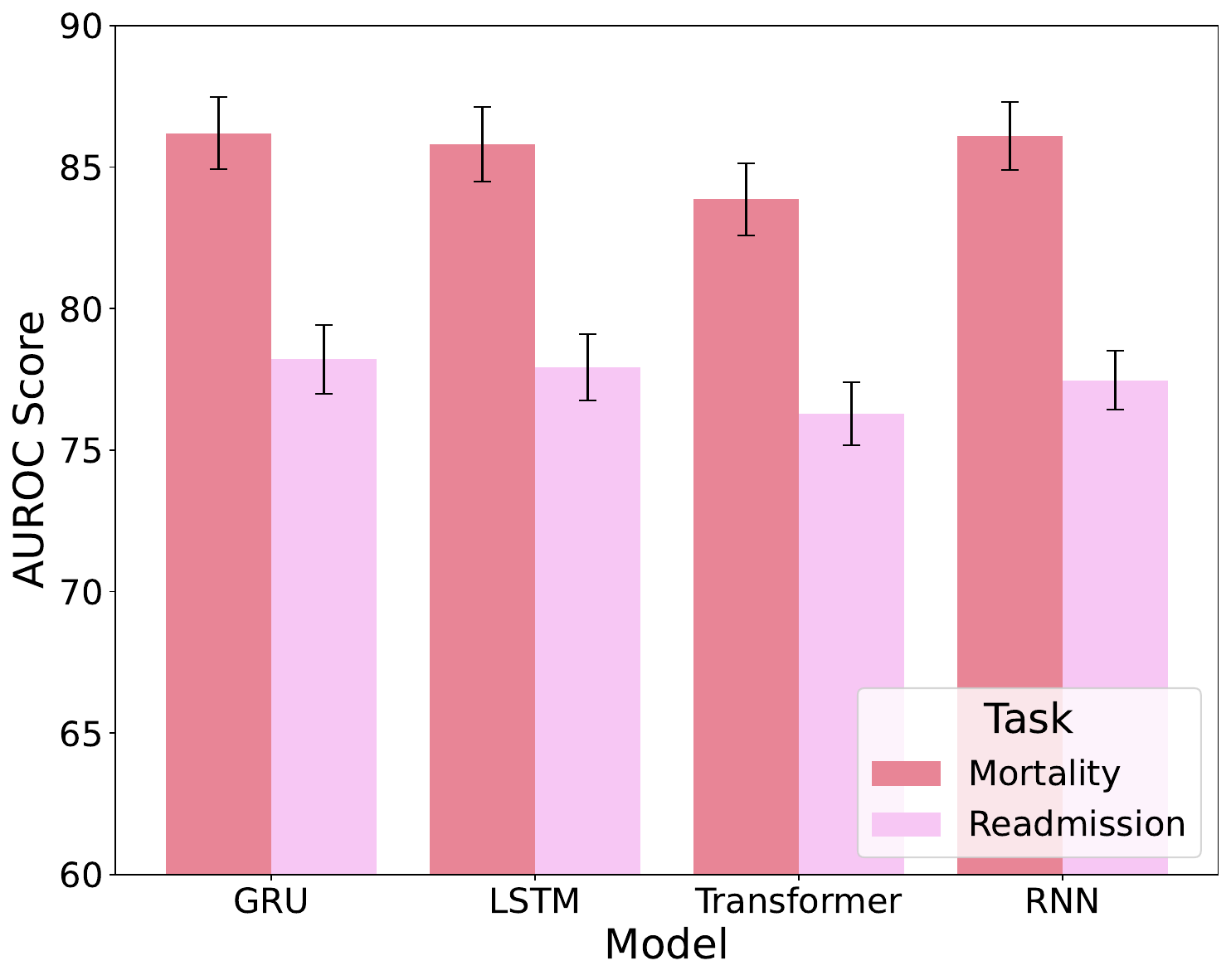}
  \label{fig:ehr_mimic3}
}
\subfigure[MIMIC-IV]{
  \includegraphics[width=0.45\linewidth]{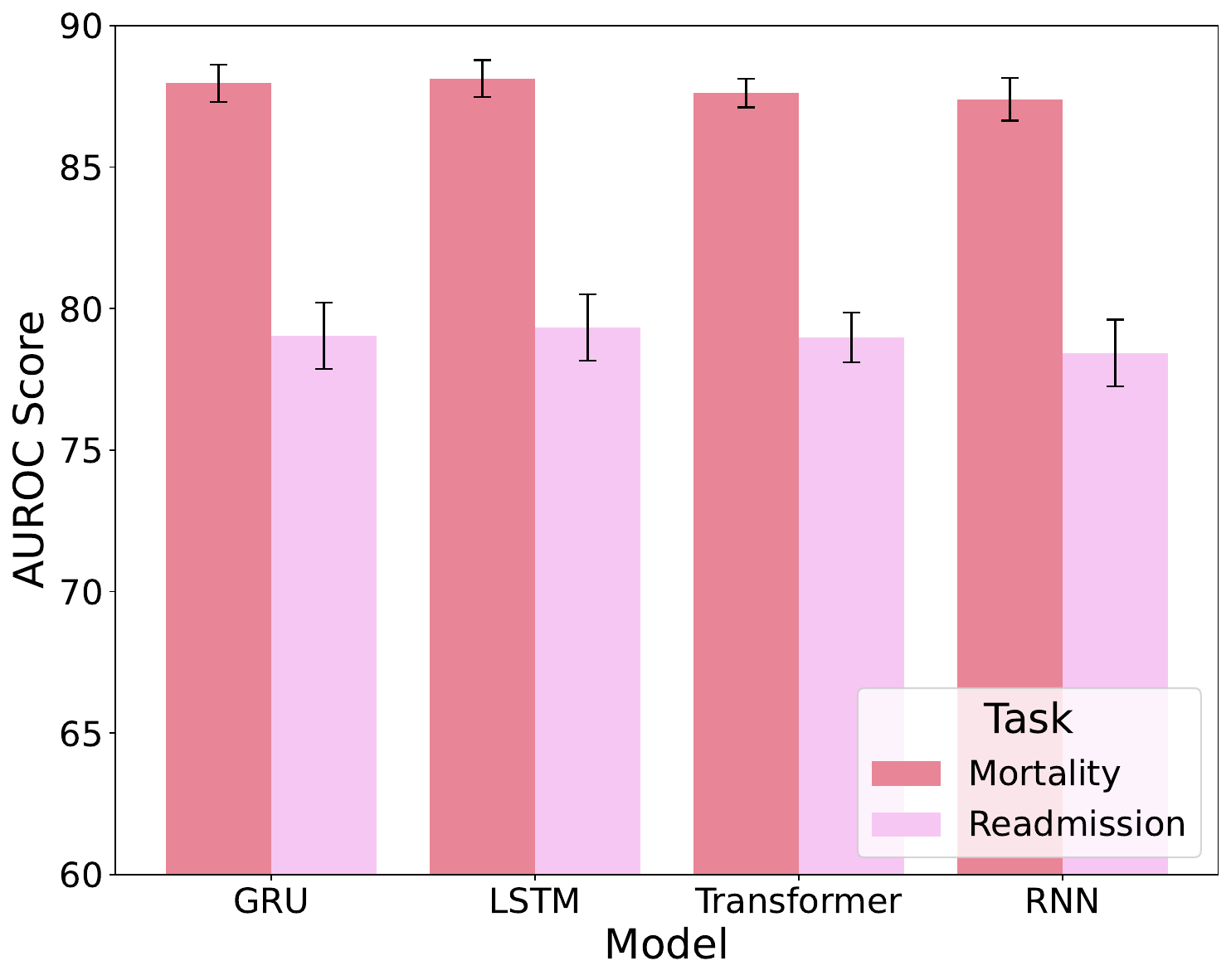}
  \label{fig:ehr_mimic4}
}
\caption{\textit{AUROC performance of four time-series encoders in in-hospital mortality prediction and 30-day readmission prediction tasks on two datasets.}}
\label{fig:ehr_roc_score}
\end{figure}

From Figure~\ref{fig:ehr_roc_score}, we compare the performance of four different time-series encoders: GRU, LSTM, Transformer, and RNN, in encoding EHR data. The evaluation focuses exclusively on time-series data inputs, excluding any text inputs, to determine which model is most effective in handling such data. The GRU model consistently performs well, therefore we have selected GRU as the backbone encoder for time-series data in \modelname{}.

\subsubsection{Comparing Different Text Fusion Approaches}

\begin{figure}[!ht]
\centering
\subfigure[MIMIC-III]{
  \includegraphics[width=0.45\linewidth]{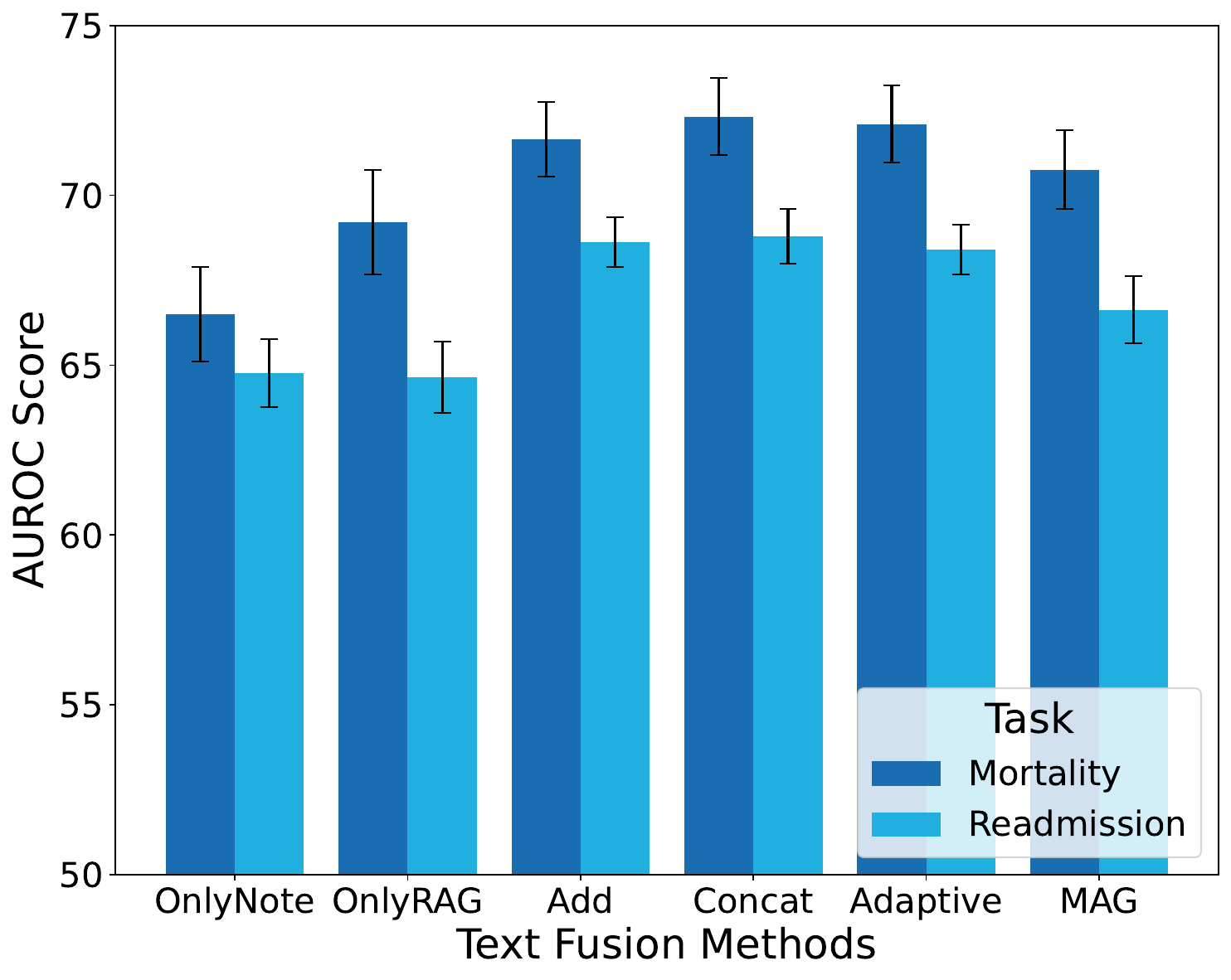}
  \label{fig:text_fusion_mimic3}
}
\subfigure[MIMIC-IV]{
  \includegraphics[width=0.45\linewidth]{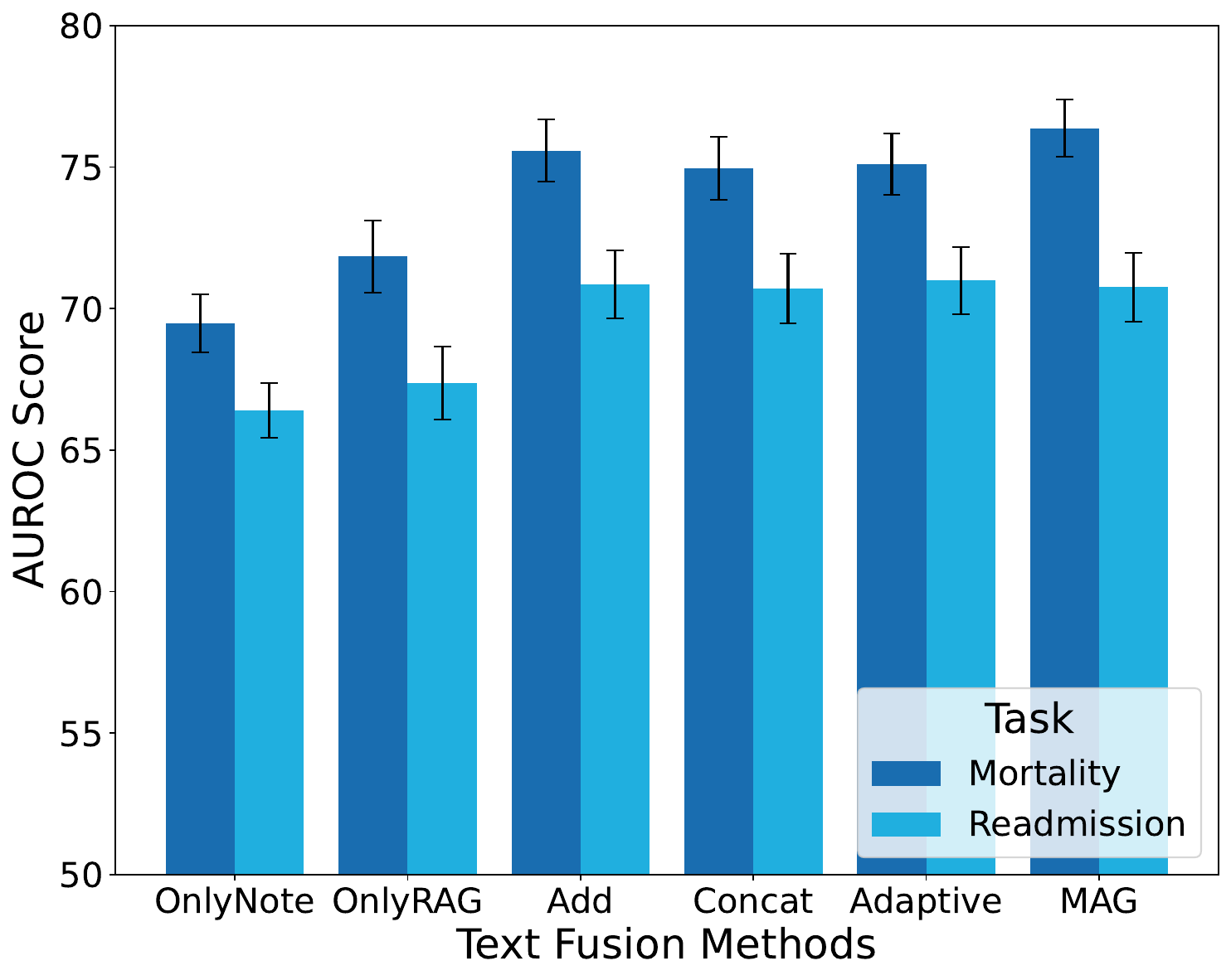}
  \label{fig:text_fusion_mimic4}
}
\caption{\textit{AUROC performance of different text fusion methods in in-hospital mortality prediction and 30-day readmission prediction tasks on two datasets.} }
\label{fig:text_fusion}
\end{figure}

From Figure~\ref{fig:text_fusion}, similar as modality fusion, we conduct the comparison for multiple text fusion approaches: note only (``OnlyNote''), summary only (``OnlyRAG''), add, concat, adaptive concat, and MAG. The evaluation focuses exclusively on text inputs with no time-series data. The concat strategy performs the best on the MIMIC-III model and shows decent performance on MIMIC-IV. Considering its simplicity, we choose concat as the text fusion method.

\subsubsection{Comparing Internal Design of Fusion Module}

\begin{figure}[!ht]
\centering
\subfigure[MIMIC-III]{
  \includegraphics[width=0.45\linewidth]{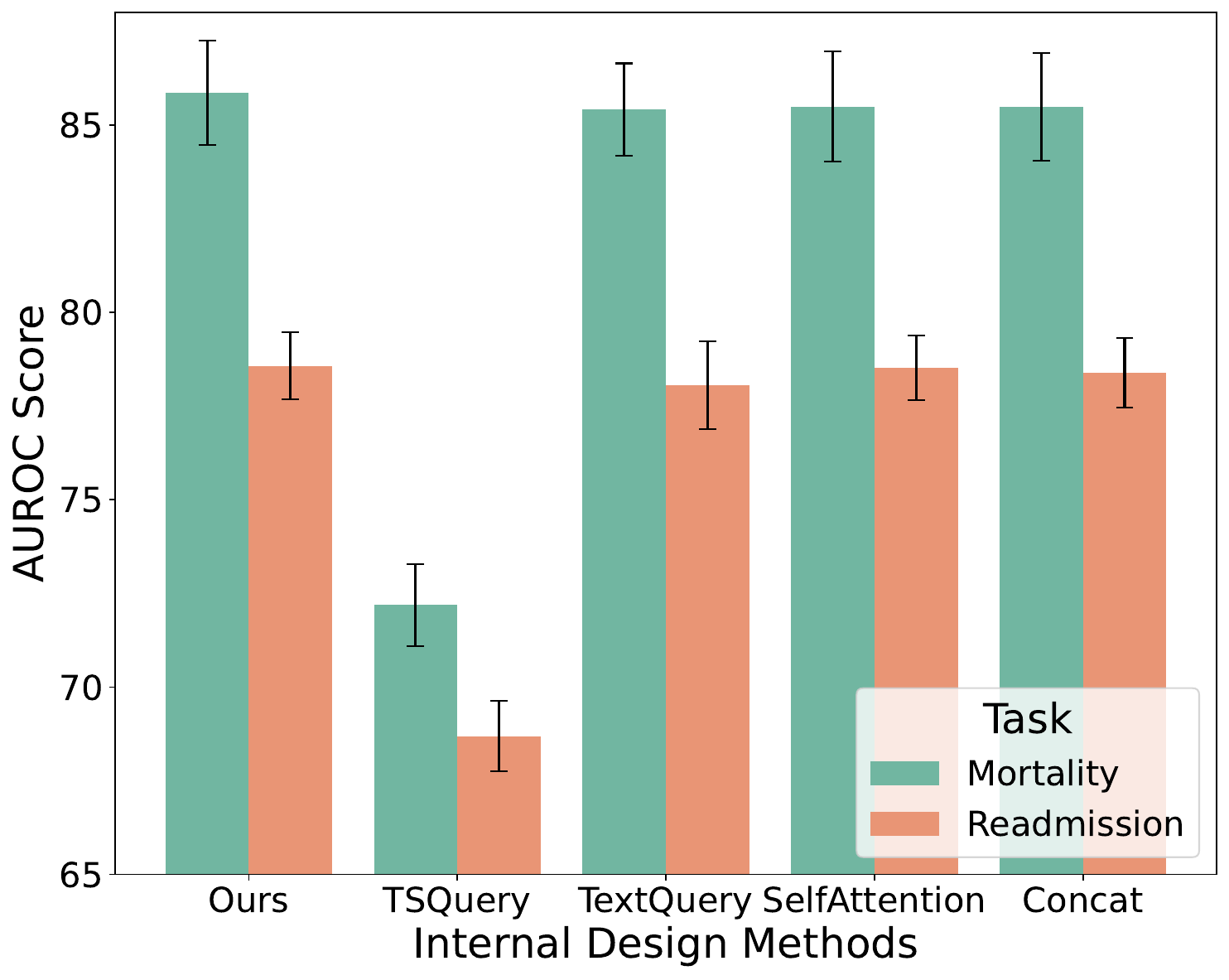}
  \label{fig:internal_fusion_mimic3}
}
\subfigure[MIMIC-IV]{
  \includegraphics[width=0.45\linewidth]{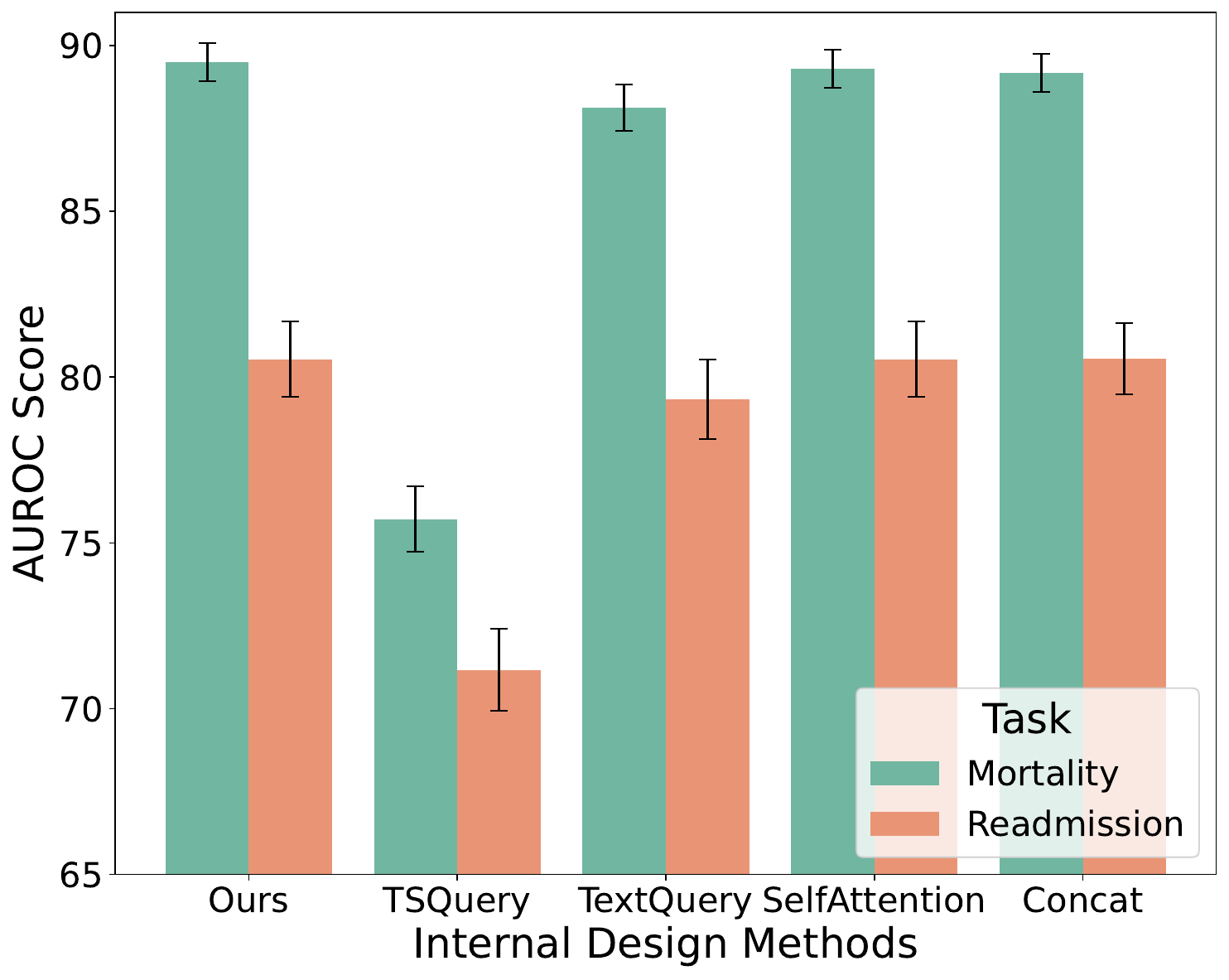}
  \label{fig:internal_fusion_mimic4}
}
\caption{\textit{AUROC performance of different internal designs of our proposed fusion module in in-hospital mortality prediction and 30-day readmission prediction tasks on two datasets.}}
\label{fig:internal_fusion}
\end{figure}

To explore in detail the role of the cross-attention mechanism for multimodal fusion in Figure~\ref{fig:fusion_module}, we provide experiments on alternative internal components in Figure~\ref{fig:internal_fusion}: ``Ours'' represents the version in Figure~\ref{fig:fusion_module}, ``TSQuery'' can be regarded as the left branch with the time-series embedding serving as the query, ``TextQuery'' as the right branch, ``SelfAttention'' replaces the cross-attention and retains the concat and projection layer, and ``Concat'' does not include any attention module. The superior performance of our final employed fusion approach demonstrates the effectiveness of the cross-modality fusion approach in a bi-directional way.

\subsection{Further Analysis}

\subsubsection{Sensitivity to Hidden Dimensions and Learning Rates}

\begin{figure}[!ht]
\centering
\subfigure[Hidden Dimensions]{
  \includegraphics[width=0.45\linewidth]{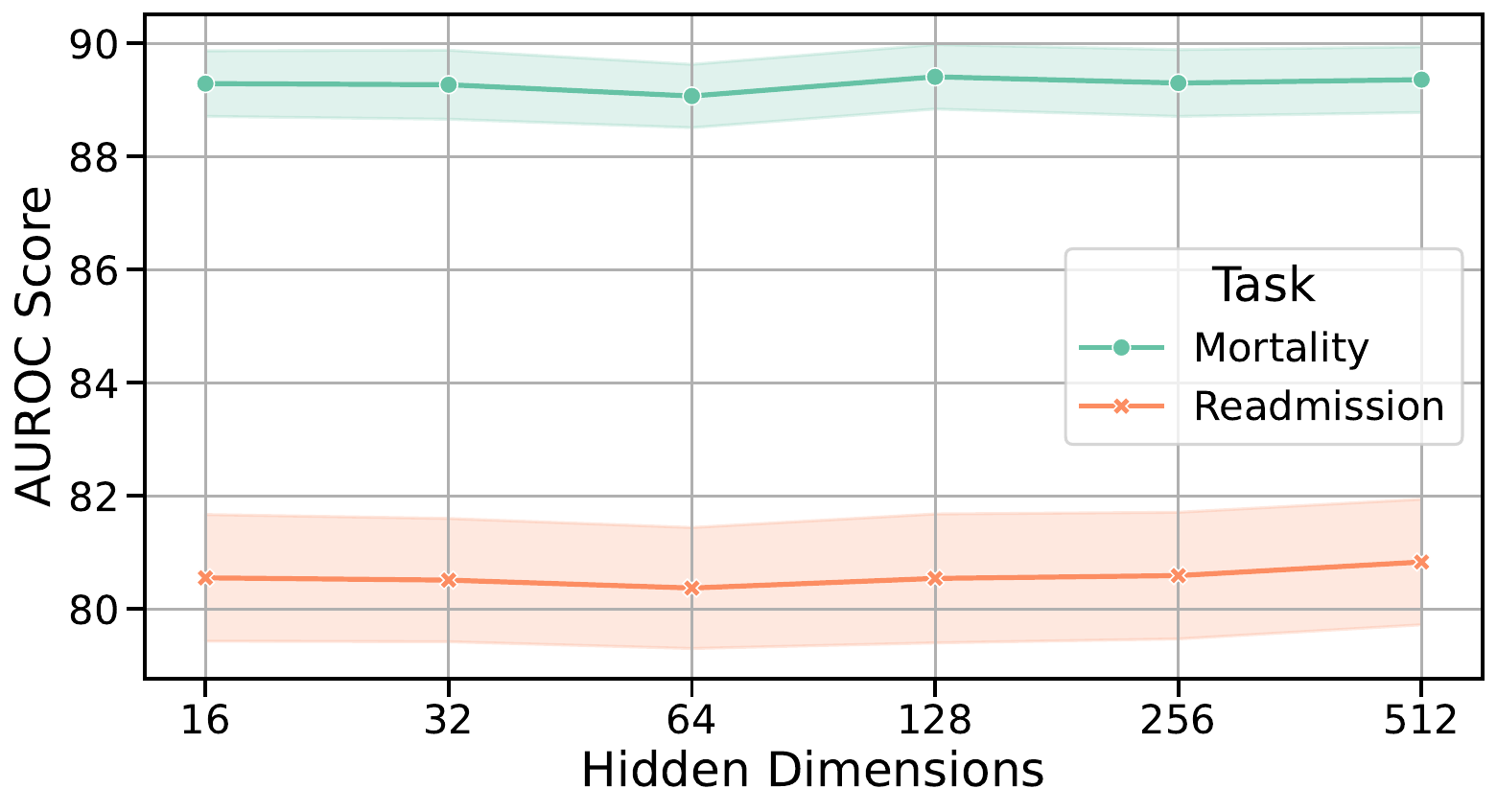}
  \label{fig:hid_sensitivity}
}
\subfigure[Learning Rates]{
  \includegraphics[width=0.45\linewidth]{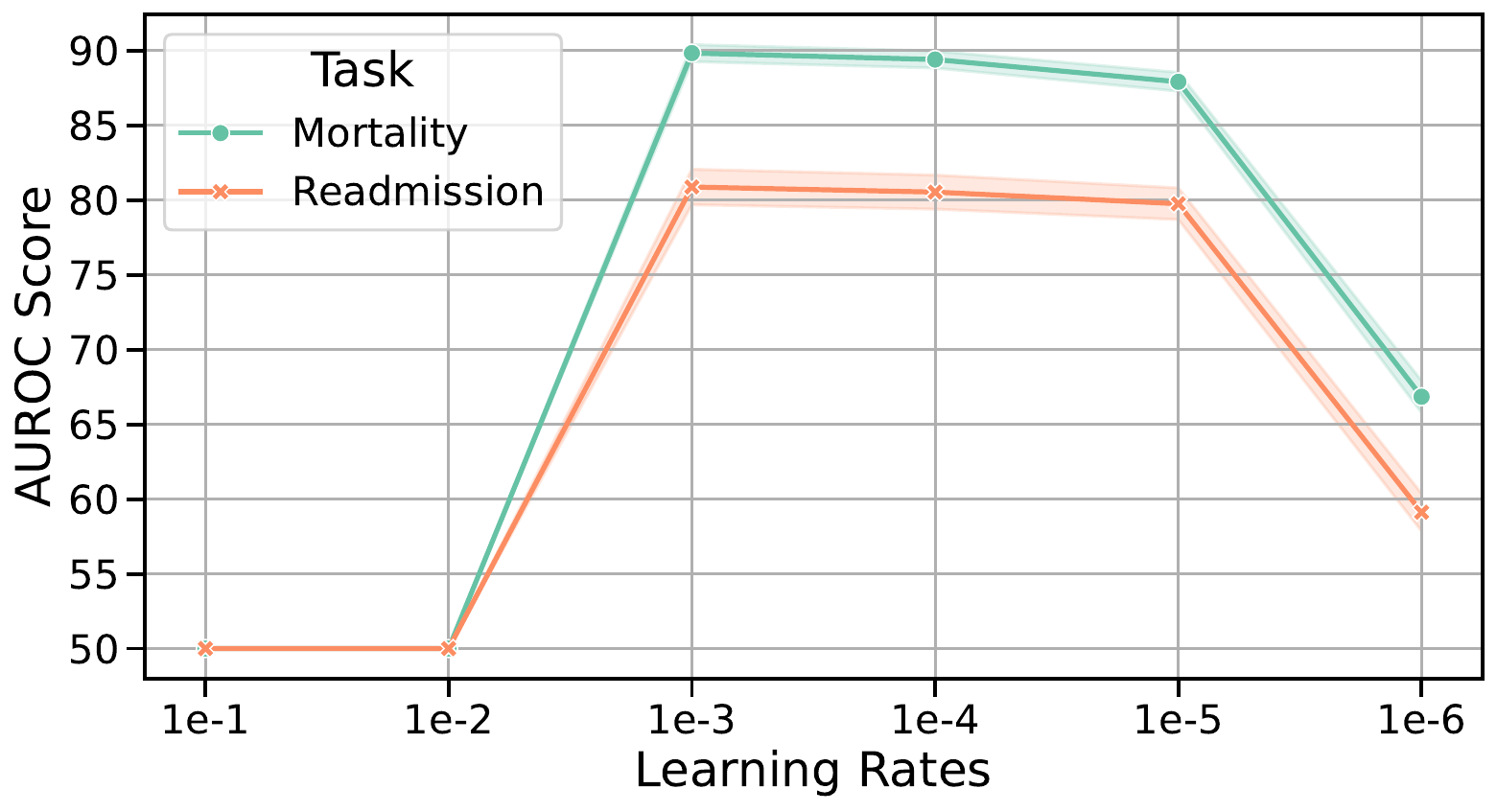}
  \label{fig:lr_sensitivity}
}
\caption{\textit{AUROC performance to various hidden dimensions (left) and learning rates (right) in in-hospital mortality and 30-day readmission prediction tasks on MIMIC-IV.} }
\label{fig:sensitivity}
\end{figure}

To assess the sensitivity of our \modelname{} framework to different hidden dimensions and learning rates, we conducted experiments on the MIMIC-III and MIMIC-IV datasets (Figure~\ref{fig:sensitivity}). The results indicate that a hidden dimension of 128 and a learning rate of 1e-3 yield the best performance. The minimal variations across different settings demonstrate \modelname{}'s low sensitivity to these hyperparameters.

\subsubsection{Robustness to Data Sparsity}

\begin{figure}[!ht]
\centering
\subfigure[Mortality]{
  \includegraphics[width=0.45\linewidth]{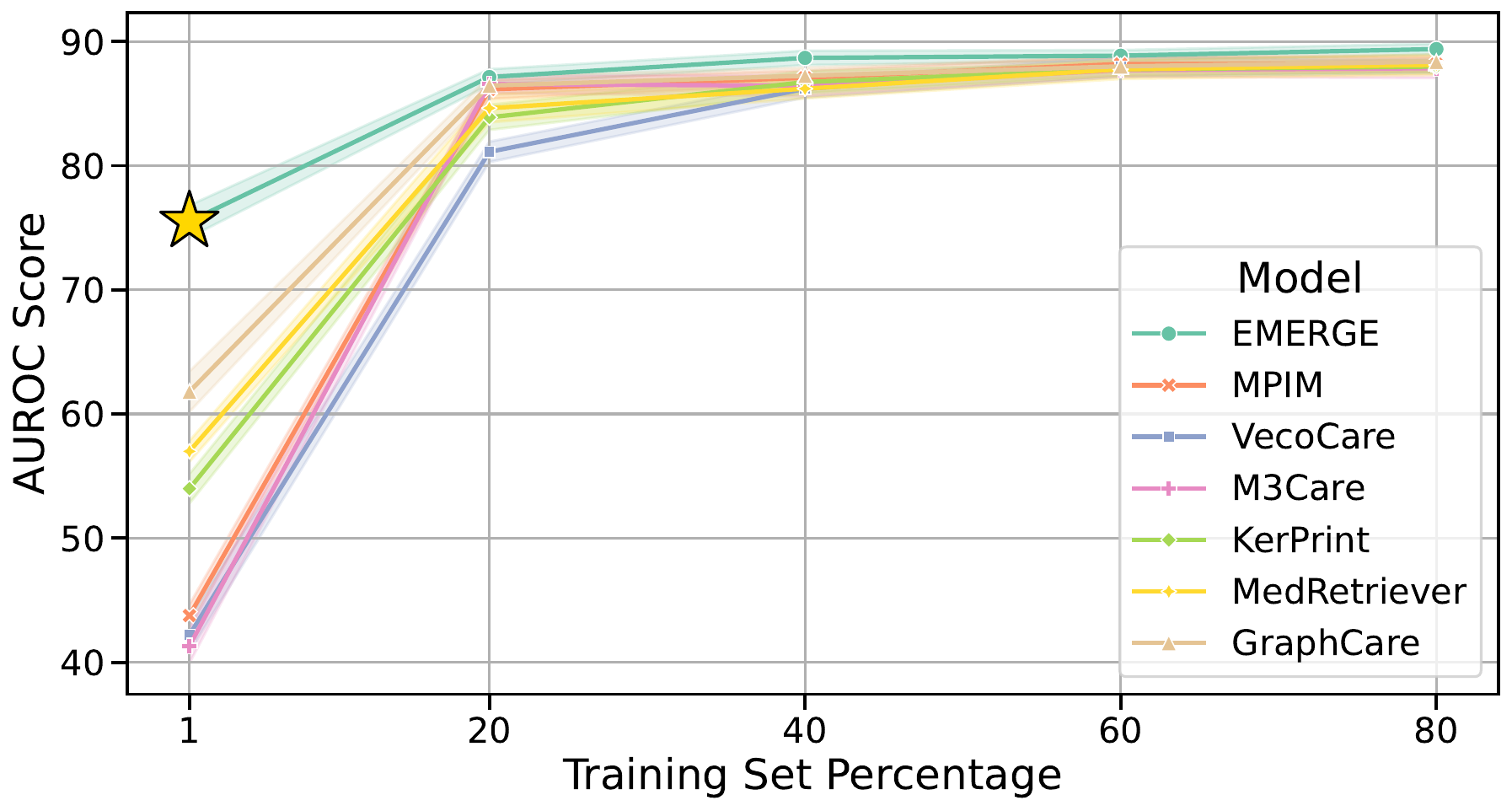}
  \label{fig:mortality_sparsity}
}
\subfigure[Readmission]{
  \includegraphics[width=0.45\linewidth]{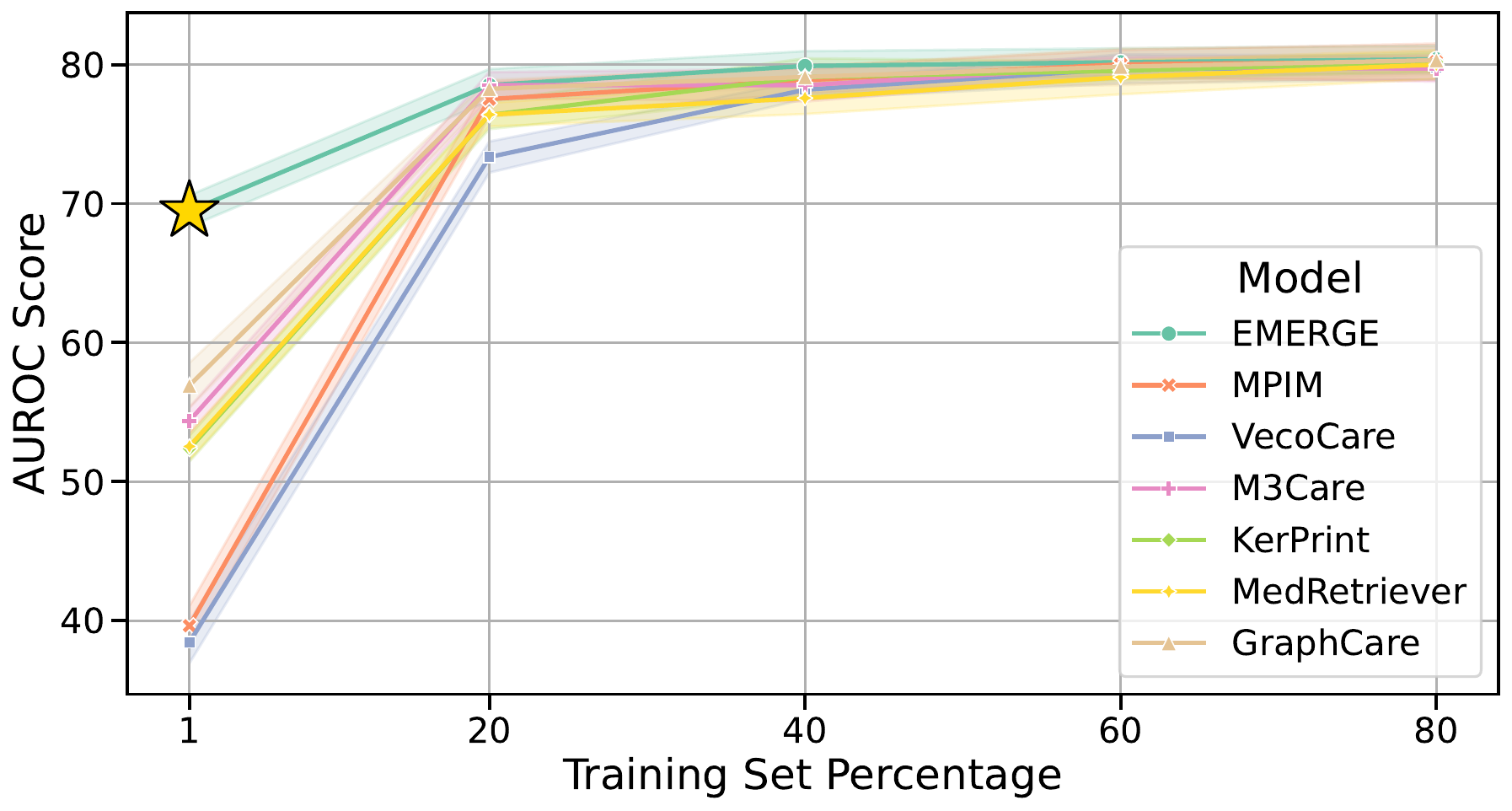}
  \label{fig:readmission_sparsity}
}
\caption{\textit{AUROC performance across 5 Training Set Percentage in in-hospital mortality prediction (left) and 30-day readmission prediction (right) task on MIMIC-IV.} }
\label{fig:sparsity}
\end{figure}

To evaluate the robustness of our \modelname{} framework against data sparsity, we conduct experiments using 1\%, 20\%, 40\%, 60\%, and 80\% of the training set. As depicted in Figure~\ref{fig:sparsity}, \modelname{} shows remarkable resilience, especially with only 1\% (less than 150) of the training samples. This robustness is crucial in clinical settings where data collection is often challenging, making \modelname{} valuable for clinical practice.

\section{Conclusions}

In this work, we propose \modelname{}, an RAG-driven multimodal EHR data representation learning framework that incorporates time-series EHR, clinical notes data, and an external knowledge graph for healthcare prediction. The \modelname{} framework comprehensively leverages LLMs' semantic reasoning ability, long-context encoding capacity, and the medical context of the knowledge graph. The \modelname{} framework achieves superior performance on two real-world datasets' in-hospital mortality and 30-day readmission tasks against the latest baseline models. Extensive experiments showcase \modelname{}'s effectiveness and robustness to data sparsity. \modelname{} marks a step towards more effective utilization of multimodal EHR data in healthcare, offering a potent solution to enhance clinical representations with external knowledge and LLMs.

\section*{Ethical Statement}

This study, involving the analysis of de-identified Electronic Health Records (EHR) from the MIMIC-III and MIMIC-IV datasets, upholds high ethical standards. It should be noted that in our use of the online API of the LLM to generate patient summaries, the content in the prompts is derived from publicly accessible knowledge graphs and only includes feature names from the MIMIC dataset. Therefore, privacy concerns are limited. Overall, our methodology aims to minimize harm and ensure unbiased, equitable findings, reflecting the complex nature of medical data. We rigorously adhere to these ethical values throughout our research.

\begin{acks}
This work was supported by the Defense Industrial Technology Development Program under Grant JCKY2021601B104.
\end{acks}

\clearpage
\newpage
\balance
\bibliographystyle{ACM-Reference-Format}
\bibliography{ref}

\end{document}